%% file: MainPaper.tex
\documentclass[twoside]{article}
\usepackage[accepted]{aistats2024}
\usepackage[dvipsnames,table,xcdraw]{xcolor}
\usepackage{graphicx}
\usepackage{subcaption}
\usepackage{svg}
\usepackage{adjustbox}
\usepackage{hyperref}
\usepackage{cite}
\usepackage{multirow}
\usepackage{float}
\usepackage{caption}
\usepackage{amsmath,amssymb,amsfonts,amsthm,clrscode3e,enumerate}
\usepackage[ruled,noend]{algorithm2e}
\usepackage{algpseudocode}
\usepackage{graphicx}
\usepackage{marginnote}
\usepackage{textcomp}
\usepackage{changepage,threeparttable}
\usepackage{booktabs}
\usepackage{xcolor}
\usepackage{tikz}
\usepackage{csquotes}
\usepackage{appendix}
\providecommand{\enquote}{\emph}
\usetikzlibrary{matrix}

\SetCommentSty{mycommfont}
\theoremstyle{definition}
\newtheorem{defn}{Definition}
\newtheorem{thm}{Theorem}
\newtheorem{lem}[thm]{Lemma}
\newtheorem{cor}[thm]{Corollary}

\newtheorem*{prf}{Proof}

\def\BibTeX{{\rm B\kern-.05em{\sc i\kern-.025em b}\kern-.08em
    T\kern-.1667em\lower.7ex\hbox{E}\kern-.125emX}}
    \usepackage{comment}
    \newcommand{\kunal}[1]{{  #1}}
    
    \newcommand{\squeezeup}{\vspace{-2.5mm}}
    
\newcommand{\e}{{\varepsilon}}
\newcommand{\trace}{{\rm Tr}}

\DeclareMathOperator{\Probability}{\mathbb{P}}

\DeclareMathOperator{\Expected}{\mathbb{E}}
\newcommand{\Ex}[1]{\Expected\pbrcx{#1}}
\newcommand{\Prob}[1]{\Probability\pbrcx{#1}}
\newcommand{\R}{{\mathbb{R}}}

\newcommand{\pth}[1]{\ensuremath{\left(#1\right)}}
\newcommand{\pbrcx}[1]{\ensuremath{\left[#1\right]}}
\renewcommand{\qed}{\hfill\blacksquare}

\newcommand{\prarabdh}[1]{{#1}}
\usepackage[accepted]{aistats2024}
\begin{document}

%

%

\twocolumn[

\aistatstitle{DiffRed: Dimensionality Reduction guided by stable rank}

\aistatsauthor{ Prarabdh Shukla \And Gagan Raj Gupta \And  Kunal Dutta }

\aistatsaddress{Department of CSE,\\ Indian Institute of Technology, Bhilai ~~~~~~~\\ Bhilai, India \And   Department of CSE, \\ ~~~~Indian Institute of Technology, Bhilai \\ Bhilai, India  \And  Institute of Informatics, \\ University of Warsaw, \\ Warsaw, Poland} ]

\begin{abstract}
In this work, we propose a novel dimensionality reduction technique, \textit{DiffRed}, which first projects the data matrix, A, along first $k_1$ principal components and the residual matrix $A^{*}$ (left after subtracting its $k_1$-rank approximation) along $k_2$ Gaussian random vectors. We evaluate \emph{M1}, the distortion of mean-squared pair-wise distance, and \emph{Stress}, the normalized value of RMS of distortion of the pairwise distances. We rigorously prove that \textit{DiffRed} achieves a general upper bound of $O\left(\sqrt{\frac{1-p}{k_2}}\right)$ on \emph{Stress} and $O\left(\frac{1-p}{\sqrt{k_2*\rho(A^{*})}}\right)$ on \emph{M1} where $p$ is the fraction of variance explained by the first $k_1$ principal components and $\rho(A^{*})$ is the \textit{stable rank} of $A^{*}$.
These bounds are tighter than the currently known results for Random maps. Our extensive experiments on a variety of real-world datasets demonstrate that \textit{DiffRed} achieves near zero \emph{M1} and much lower values of \emph{Stress} as compared to the well-known dimensionality reduction techniques. In particular, \textit{DiffRed} can map a 6 million dimensional dataset to 10 dimensions with 54\% lower \emph{Stress} than PCA.
\end{abstract}

\section{Introduction}

High dimensional data is common in biological sciences, fin-tech, satellite imaging, computer vision etc. which make tasks such as machine learning, data visualization, similarity search, anomaly detection, noise removal etc. very difficult. Dimensionality reduction is a pre-processing step to obtain a low-dimensional representation while preserving its \enquote{structure} and \enquote{variation}. In this work, our focus is on the development of efficient dimensionality reduction algorithms that map $D$-dimensional data in $\mathbb{R}^D$ to $\mathbb{R}^{d}$ where $d$, the target dimension is a small number. This decreases the amount of training time and computation resources required for the above tasks. 

We consider two metrics to quantify distortion, which we aim to minimize. The first metric \emph{M1} is the distortion of mean-squared pair-wise distances. Minimizing \emph{M1} ensures that the low-dimensional representation has similar ``Energy'' or ``total variance'' as the original data. While this is important, we also need to preserve both short and long pair-wise distances for preserving importance structures such the nearest-neighbors and clusters in the data. This is accomplished by minimizing \emph{Stress}\cite{Kruskal-Stress}, the normalized value of RMS distortion of the pairwise distance by the mapping. \prarabdh{While \emph{M1} may be minimized by a simple scaling of data points, doing so may distort other metrics such as \emph{Stress}.}

Traditional dimensionality reduction techniques such as PCA, SVD, MDS \cite{HR-PCA, PCAvsRMAP} use the structure of data to determine directions along which data should be projected. One identifies the ``elbow" in a \emph{scree plot} to choose the number of principal components. Beyond this, one gets diminishing returns and is forced to either choose a large target dimension or accept high distortion. In contrast to these approaches, one can use data-agnostic Gaussian random maps ~\cite{10.1145/502512.502546} which minimize distortion of pair-wise distances. It was generally thought that to guarantee low distortion, a large number of target dimensions are required by Gaussian random maps. In a recent work, \cite{Bartal1} obtained bounds of the form $O\left(\frac{1}{\sqrt{d}}\right)$ on \emph{Stress} when using Gaussian random maps of any arbitrary dimension $d$. They also demonstrated that PCA can produce an embedding with \emph{Stress} value being far from optimum and random maps can achieve better performance. 

\prarabdh{We propose a novel approach to dimensionality reduction that uses the \textit{stable rank} (Def in Sec \ref{sec:prob-form}) of the data. The stable rank of a dataset gives an idea of directional spread in the data. It is always greater than 1 and less than the actual rank of the data. If the data is spread along various directions, its stable rank will be high, and if it is concentrated along a few directions only, then the stable rank will be low (refer Figure \ref{fig:stable-rank-intuition}). Intuitively, for datasets with low stable rank, PCA is more effective. Our findings reveal a fresh perspective: Random Maps are more effective for high stable rank datasets, as opposed to the conventional belief that Random Maps are data-agnostic.}

\input{figure_files/stable-rank-intuition}

\prarabdh{In this work}, we prove rigorously using Hanson Wright inequality \cite{HWineq}, that \emph{M1} can be bounded by $O\left(\frac{1}{\sqrt{d\rho(A)}}\right)$ where $\rho(A)$ is the stable rank ( Def in Sec 3) of the data matrix $A$. Thus, if stable rank is high, we can guarantee low distortion for small values of $d$. Empirically, we observe a similar behavior with respect to \emph{Stress}. Since all input data matrices may not have high stable rank, we subtract the best k-rank approximation of input data matrix and obtain the residual matrix $A^{*}$. Empirically, we observe that for most common high-dimensional datasets, $A^{*}$ has a higher stable rank than $A$ and using random maps for dimensionality reduction will minimize its distortion. \textit{DiffRed} leverages these insights and first projects the data along first $k_1$ principal components such that the fraction of variance, $p$ explained by them is high. In the next step, it projects $A^{*}$ along $k_2$ Gaussian random vectors. We make sure that these two projections lie in orthogonal subspaces, which plays a crucial role in obtaining a tighter upper bound of $O\left(\sqrt{\frac{1-p}{k_2}}\right)$ for \emph{Stress}. This gives us a good analytical trade-off between the number of principal components and random vectors used to minimize \emph{Stress} while keeping the target dimension $d=k_1+k_2$ small. \prarabdh{ We demonstrate that \textit{DiffRed} is effective on high dimensional datasets. It preserves global structure even with low target dimensions by carefully choosing $k_1$, such that the stable rank of the residual matrix is high and the theoretical bound is minimized.}

To summarize, our contributions in this paper are as follows:
\begin {itemize}

\item{We develop a new dimensionality reduction algorithm, \textit{DiffRed} that combines Principal Components with Gaussian random maps in a novel way to achieve tighter upper bounds on both \emph{M1} and \emph{Stress} metrics.}

\item{To the best of our knowledge, we are the first to have incorporated a metric involving the structure of the data matrix (Stable Rank) and impact of Monte-Carlo iterations in the bound of \emph{M1} and \emph{Stress} for \textit{DiffRed} and Random maps. This allows $d$ to be small and explains why random maps often work well in practice for high-dimensional datasets.}

\item{Fast implementation of \textit{DiffRed} and extensive experiments to demonstrate that it achieves better performance than various commonly used dimensionality reduction techniques on real-life datasets.}

\end{itemize}

\section{Related Work}
Dimensionality reduction has been studied by~\cite{10.1145/1143844.1143866,6420844,10.5555/1005332.1005335,10.5555/1005332.1005346} in the context of machine learning. 
In the broader context of metric embedding, there is a large body of work in  
diverse research areas demonstrating the practicality of various dimensionality reduction and metric embedding techniques, e.g.~\cite{1019258}.
Dimensionality reduction techniques can be broadly classified as $(i)$ linear and $(ii)$ non-linear. 

The most common linear dimensionality reduction technique is PCA (Principal Component Analysis)~\cite{PCA-Original}, although several other classical techniques such as factor analysis and multidimensional scaling, are also used~\cite{10.2307/1412107,Torgerson1952}. However, these linear techniques are not very good at handling non-linear data, e.g. when the data is lying on a 
low-dimensional manifold in a high-dimensional ambient space -- often referred to as the \emph{manifold hypothesis}~\cite{ea1cd7d492fc4f829505bf918b3bcf7d}. 

In contrast, non-linear techniques such as Kernel PCA, Isomap, Diffusion maps, or Locally Linear Embedding, etc. can be quite effective at handling particular types of non-linear data, such as convex or Gaussian data, and are being used more and more in recent applications~\cite{vandermaaten2009dimensionality}. However, in general, the technique used needs to be tailored to the application, as certain maps can be quite bad for certain types of datasets.

In this scenario, the method of random projections~\cite{10.1145/502512.502546} is a linear dimensionality reduction technique, which has the advantages of genericity, low computational complexity, low memory requirement, and ability to handle some degree of non-linearity, e.g. data lying in low-dimensional 
manifolds -- in contrast to PCA which, for high-dimensional data, requires significant computational time and memory, and cannot handle non-linear data. Various time-efficient randomized variants of PCA and SVD have been proposed, such as \cite{halko2010finding, feng2018fast}. Similarly, faster variants of the Random Map have been proposed \cite{fastJL}. Recently, \cite{fandina2022fast} have presented a fresh analysis of the Fast JL transform, showing an improvement in embedding time. \cite{Schmidt2018Stable} is one of the few comparative studies involving PCA and random projections. 

\kunal{The notion of \emph{stable rank} (or numerical rank) of a matrix is a robust version of the rank of a matrix, and is not affected significantly by very small singular values. It was first introduced in~\cite{10.1145/1255443.1255449} who used it to obtain low-rank approximations of matrices. Since then it has found several applications in numerical linear algebra e.g.~\cite{NEURIPS2019_1625abb8, cohen_et_al:LIPIcs:2016:6278,DBLP:conf/colt/Kasiviswanathan18}. However to the best of our knowledge, it has not yet been used to obtain stronger dimensionality reduction bounds. Moreover, in the known applications of stable rank, it is advantageous to have low stable rank, e.g. to obtain low rank approximations of matrices. On the other hand, our application utilizes \emph{high} stable rank, which gives a stronger concentration bound for the mapped vectors, allowing us to choose a lower target dimension.}

\prarabdh{\emph{Stress} as a metric has been used in a variety of applications such as MDS \cite{Kruskal-Stress}, psychology \cite{Borg2005} and also surface matching \cite{Bronstein2006GeneralizedMS} which is applied to 3D face recognition and medical imaging. Various quantitative studies of dimensionality reduction such as \cite{Espadoto2019TowardAQ, yin-stress, liu-stress} have also considered \emph{Stress} to be an important distortion metric to measure projection quality.}


Stochastic embedding methods such as T-SNE \cite{vanDerMaaten2008} are also popular for visualization of datasets. However, they can cause large distortion and are 
rarely used for tasks such as machine learning, similarity search, anomaly detection, noise removal etc. UMAP \cite{mcinnes2020umap} is another useful visualization technique that performs manifold learning. Unlike T-SNE, it has no restriction on the target dimension.

\prarabdh{In recent years, \cite{Espadoto2019TowardAQ} is the most comprehensive survey of Dimensionality Reduction techniques. They work with 18 datasets, 44 techniques, and 7 quality metrics to create a projection assessment benchmark that helps answer which dimensionality reduction algorithm applies to a given context. In our experiments, we compare \textit{DiffRed} to the best techniques reported in their survey.}

\section{Problem Formulation}\label{sec:prob-form}
Let us now formally define the problem of dimensionality reduction of a data matrix $A$ to obtain embedding matrix $\Tilde{A}$ while minimizing \emph{M1} and \emph{Stress}.
\begin{defn}[Data Matrix]
A matrix in $\mathbb{R}^{n\times D}$ whose rows are $n$ points $\mathbf{x}_1^\top,\dots,\mathbf{x}_n^\top$ in $\mathbb{R}^D$ is called a Data Matrix and is denoted by $A$. Without loss of generality, we will assume that $A$ has rows with mean zero and unit variance\footnote{This assumption will help us in proving Lemma \ref{data-difference}}. 
\end{defn}
 
\begin{defn}[Embedding Matrix]
Given a data matrix $A\in \mathbb{R}^{n\times D}$, its corresponding embedding matrix $\Tilde{A}\in \mathbb{R}^{n\times d}$ is a matrix whose rows $\mathbf{\Tilde{x}}^\top_1,\dots,\mathbf{\Tilde{x}}^\top_n$ are embeddings of the rows of $A$ onto $\mathbb{R}^d$. 

\prarabdh{From now on, $d$ shall denote the target dimension and $D$ shall denote the original dimension unless specified otherwise.} 
\end{defn}
\begin{defn}[Stable Rank]
For a given matrix $A$, let $\sigma_1,\sigma_2,\dots$ be the singular values ordered from the highest to the lowest in magnitude. Then, the stable rank $\rho(A)$ of $A$ is defined as 
$$\rho(A)=\frac{\sum_{i=1}^{rank(A)}\sigma_i^2}{\sigma_1^2}$$
\end{defn}
\begin{defn}[\emph{M1} Distortion]

For data matrix $A\in \mathbb{R}^{n\times D}$ (whose rows $\mathbf{x}_1^\top,\dots,\mathbf{x}_n^\top \in \mathbb{R}^D$ are the data points) and its corresponding embedding matrix $\Tilde{A} \in \mathbb{R}^{n\times d}$ (whose rows $\mathbf{\Tilde{x}}^\top_1,\dots,\mathbf{\Tilde{x}}^\top_n$  in $\mathbb{R}^d$ are the low dimensional embeddings), the \emph{M1} distortion ($\Lambda_{M_1}$) is given by:
$$\Lambda_{M1}(A,\Tilde{A})=\left| 1- \frac{||\Tilde{A}||_F^2}{||A||_F^2}\right |= \left |1- \frac{\sum_{i=1}^n ||\mathbf{\Tilde{x}}||_2^2}{\sum_{i=1}^n ||\mathbf{x}||_2^2}\right |$$
\end{defn}
\begin{defn}[\emph{Stress}]
For a set of points $\mathbf{x}_1,\dots,\mathbf{x}_n$ in D-dimensional space $\mathbb{R}^D$ and their respective low dimensional embeddings, $\mathbf{\Tilde{x}}^\top_1,\dots,\mathbf{\Tilde{x}}^\top_n$  in $\mathbb{R}^d$, we define the \emph{Stress} $\Lambda_S$ as:
$$\Lambda_S=\left(\frac{\sum_{i,j}(||\mathbf{d}_{ij}||-||\mathbf{\Tilde{d}}_{ij}||)^2}{\sum_{i,j}||\mathbf{d}_{ij}||^2}\right)^\frac{1}{2}$$
where, $\mathbf{d}_{ij}=\mathbf{x}_i-\mathbf{x}_j$ and $\mathbf{\Tilde{d}}_{ij}=\mathbf{\Tilde{x}}_i-\mathbf{\Tilde{x}}_j$
\end{defn}
\begin{defn}[p]
$p=\frac{\sum_{i=1}^{k_1}\sigma_i^2}{\sum_{i=1}^r\sigma_i^2}$ represents the fraction of variance explained by $k_1$ principal components of $A$.
\end{defn}

\section{\textit{DiffRed} Algorithm and Its Analysis}\label{sec:dr-algo}
In this section, we formally describe the \textit{DiffRed} algorithm, which uses a combination of principal components and Gaussian random maps to provide provable low distortion. In the pseduocode below, SVD has been employed for PCA and k-rank approximation. Alternatively, eigen-decomposition can also be used. 

\begin{algorithm}
\DontPrintSemicolon
\SetNoFillComment
\setlength{\abovedisplayskip}{3pt}
\setlength{\belowdisplayskip}{3pt}
\setlength{\abovedisplayshortskip}{3pt}
\setlength{\belowdisplayshortskip}{3pt}
\KwIn{$A$, $k_1$, $k_2$, $\eta$}
\caption{\textit{DiffRed} Algorithm}
\label{DiffRed-algo}
    compute\footnotemark SVD $A=U\Sigma V^\top$ \label{fast-computations}\\
    compute $A_{k_1} \leftarrow \sum_{i=1}^{k_1}\sigma_i\mathbf{u}_i\mathbf{v}_i^\top$ and $A^{*} \leftarrow A - A_{k_1}$ \\
    Let $V_{k_1}$ be the matrix with the $k_1$ leftmost columns of $V$\\
    $Z\leftarrow AV_{k_1}$ \tcp*{Project $A$ along $V_{k_1}$}
    Initialize $min=\infty$ \\
    Initialize $T, T_{min} \in \mathbb{R}^{n\times k_2}$ \\ 
    \tcp*{$\eta$ Monte Carlo iterations}
    \For{$i=0, \cdots, \eta$}{
        Sample $G\in \mathbb{R}^{D\times k_2}$ where  $G_{ij}\sim\mathcal{N}\left(0,1\right)$ i.i.d. \\
    $G\leftarrow \frac{1}{\sqrt{k_2}}G$\\
        $T \leftarrow A^* G$\\
        \If{$\Lambda_{M_1}(A^*,T) < min$}{$T_{min}\leftarrow T$}
            
    }
    $R\leftarrow T_{min}$\\ \tcp*{$T_{min}$ is the projection with least $\Lambda_{M_1}$}
    $\Tilde{A} \leftarrow [Z|R]$\\
    \Return $\Tilde{A}$
\end{algorithm}
\footnotetext{When $U$ and $V$ both are extremely large, a custom power iteration algorithm may be used to calculate only the top $k_1$ singular vectors and singular values}

\begin{figure}
\captionsetup{font=small}
    \centering
\tikzset{every picture/.style={line width=0.75pt}} 
\begin{tikzpicture}[x=0.75pt,y=0.75pt,yscale=-0.75,xscale=0.75]

\draw    (246,173) -- (393,100.88) ;
\draw [shift={(394.8,100)}, rotate = 153.87] [color={rgb, 255:red, 0; green, 0; blue, 0 }  ][line width=0.75]    (10.93,-3.29) .. controls (6.95,-1.4) and (3.31,-0.3) .. (0,0) .. controls (3.31,0.3) and (6.95,1.4) .. (10.93,3.29)   ;
\draw   (192,173) .. controls (192,143.18) and (216.18,119) .. (246,119) .. controls (275.82,119) and (300,143.18) .. (300,173) .. controls (300,202.82) and (275.82,227) .. (246,227) .. controls (216.18,227) and (192,202.82) .. (192,173) -- cycle ;
\draw    (246,173) -- (276.66,128.65) ;
\draw [shift={(277.8,127)}, rotate = 124.66] [color={rgb, 255:red, 0; green, 0; blue, 0 }  ][line width=0.75]    (10.93,-3.29) .. controls (6.95,-1.4) and (3.31,-0.3) .. (0,0) .. controls (3.31,0.3) and (6.95,1.4) .. (10.93,3.29)   ;
\draw    (277.8,127) -- (392.85,100.45) ;
\draw [shift={(394.8,100)}, rotate = 167.01] [color={rgb, 255:red, 0; green, 0; blue, 0 }  ][line width=0.75]    (10.93,-3.29) .. controls (6.95,-1.4) and (3.31,-0.3) .. (0,0) .. controls (3.31,0.3) and (6.95,1.4) .. (10.93,3.29)   ;
\draw    (277.8,127) -- (257.26,72.87) ;
\draw [shift={(256.55,71)}, rotate = 69.22] [color={rgb, 255:red, 0; green, 0; blue, 0 }  ][line width=0.75]    (10.93,-3.29) .. controls (6.95,-1.4) and (3.31,-0.3) .. (0,0) .. controls (3.31,0.3) and (6.95,1.4) .. (10.93,3.29)   ;
\draw    (246,173) -- (256.34,72.99) ;
\draw [shift={(256.55,71)}, rotate = 95.9] [color={rgb, 255:red, 0; green, 0; blue, 0 }  ][line width=0.75]    (10.93,-3.29) .. controls (6.95,-1.4) and (3.31,-0.3) .. (0,0) .. controls (3.31,0.3) and (6.95,1.4) .. (10.93,3.29)   ;

\draw (328,136.4) node [anchor=north west][inner sep=0.75pt]    {$\mathbf{x}$};
\draw (271,137.4) node [anchor=north west][inner sep=0.75pt]    {$\mathbf{z}$};
\draw (310,95.4) node [anchor=north west][inner sep=0.75pt]    {$\mathbf{y}$};
\draw (279,85.4) node [anchor=north west][inner sep=0.75pt]    {$\mathbf{r}$};
\draw (237,90.4) node [anchor=north west][inner sep=0.75pt]    {$\mathbf{\tilde{x}}$};
\draw (237,194.4) node [anchor=north west][inner sep=0.75pt]    {$S_{k_1}$};

\end{tikzpicture}

    \caption{\textit{DiffRed} algorithm maps vector $\mathbf{x} \in \mathbb{R}^D $ to $\mathbf{\tilde{x}} \in \mathbb{R}^{k_1+k_2}$ while preserving its component  $\mathbf{z}$ in the best-fit-subspace $S_{k_1}$. $\mathbf{r}$ and $\mathbf{y}$ are orthogonal to $\mathbf{z}$.  }
    \label{fig:DiffRed}
\end{figure}
Each vector $\mathbf{x}\in \mathbb{R}^D$ can be written as the following sum: $\mathbf{x}=\mathbf{z}+\mathbf{y}$.
Here, $\mathbf{z} \in S_{k_1}=\proc{span}(\mathbf{v}_1,\dots,\mathbf{v}_{k_1})$ where the v's are the $k_1$ principal components of the data matrix (which span the row space). $\mathbf{y}$ lies in the residual subspace $\mathbb{R}^D \setminus S_{k_1}$ and is orthogonal to $\mathbf{z}$ by definition. $\mathbf{z}$ is fully preserved during dimensionality reduction chosen by \textit{DiffRed}. Only $\mathbf{y}$ undergoes a projection via random map to give $\mathbf{r}$. Finally, our embedded vector becomes $\mathbf{\tilde{x}}=\mathbf{z}+\mathbf{r}$.
Our claim is that the square of the difference between length (norm) of $\mathbf{x}$ and $\mathbf{\tilde{x}}$ is less than that of between $\mathbf{r}$ and $\mathbf{y}$, i.e., $(|\mathbf{x}|-|\mathbf{\tilde{x}}|)^2\leq (|\mathbf{y}|-|\mathbf{r}|)^2$. PCA and its variants attempt to preserve only $\mathbf{z}$ while neglecting $\mathbf{y}$ completely. \textit{DiffRed} solves this problem elegantly. Increasing $k_1$ allows us to preserve \enquote{longer} $\mathbf{z}$ while increasing $k_2$ reduces the distortion of $\mathbf{y}$. In the proofs below, these insights are extended to the full data matrix, $A$.

Lemma \ref{l:rand-map-avg-dist} presents a tighter upper bound on \emph{M1} for Gaussian random projections using the notion of stable rank and Theorem \ref{M1Bound} does the same for \textit{DiffRed}. Corollary \ref{cor:M1Bound} 
analyzes the importance of performing Monte Carle iterations in \textit{DiffRed}. Then, we state a recent result on bounding \emph{Stress} in Theorem \ref{bartal} \cite{Bartal1}. Theorem \ref{stressBound} proves a tighter bound on \emph{Stress} achieved by \textit{DiffRed}.

\begin{lem}
         \label{l:rand-map-avg-dist}
              There exists a constant $c_1>0$, such that given a random matrix $G$ as defined in the \textit{DiffRed} Algorithm \ref{DiffRed-algo} and a data matrix $A$, 
            for all $d \leq D$ and all $\e \in [0,1]$
          \[ \Prob{|\|AG\|_F^2-\|A\|_F^2|\geq \e\cdot \|A\|_F^2} \leq 2.\exp\pth{-c_1\e^2d \rho_A}.\]
         \end{lem}
\begin{thm}[\emph{M1} Distortion Bound]\label{M1Bound}
Given a data matrix $A\in \mathbb{R}^{n\times D}$ \kunal{and non-negative integers} $k_1$ and $k_2$, let the application of the \textit{DiffRed} algorithm on $A$ with target dimensions $k_1$ and $k_2$ return the embedding matrix $\Tilde{A}\in \mathbb{R}^{n\times d}$ where $d=k_1+k_2$. Then, 

$$\Prob{\Lambda_{M1}(A)\geq \e}\leq 2e^{(-\frac{c_1\e^2k_2\rho(A^{*})}{(1-p)^2})} $$
where $c_1>0$ is a constant.  
\end{thm}
The proof of Theorem \ref{M1Bound} is provided in the supplementary material.

To minimize $\Prob{\Lambda_{M1}\geq \e}$ (failure probability), the argument in the exponent above needs to be large. This means we can achieve an $\e$ of the order of $\frac{(1-p)}{\sqrt{k_2\rho(A^{*})}}$. Performing Monte Carlo iterations reduces the failure probability considerably and \emph{M1} can be minimized.
The following corollary is a direct consequence of Theorem \ref{M1Bound}
\begin{cor}[\emph{M1} bound for RMap]\label{cor:RMapM1Bound}
From Theorem \ref{M1Bound}, for the case of pure Random Maps($p=0$,$k_1=0$, $k_2=d$), we have the following bound:
$$\Prob{\Lambda_{M1}(A)\geq \e}\leq 2e^{(-c_1\e^2d\rho(A))} $$

\end{cor}
\begin{cor}[\emph{M1} Distortion, Monte Carlo Version]\label{cor:M1Bound}
Given a data matrix $A\in \mathbb{R}^{n\times D}$,$k_1$ and $k_2$, and given $\eta>0$, let the application of the \textit{DiffRed} algorithm on $A$ with target dimensions $k_1$ and $k_2$ return the embedding matrix $\Tilde{A}\in \mathbb{R}^{n\times d}$ where $d=k_1+k_2$. Then, 
the probability that in $\eta$ Monte Carlo iterations,
\begin{eqnarray*}
    \Prob{\min\{\Lambda_{M1}(A)\}\geq \e} \leq & \\ \delta_0^{\eta} \leq \\  \exp\pth{-\eta\pth{\frac{ c_1\e^2k_2\rho(A^{*})}{(1-p)^2}-\ln 2}} 
\end{eqnarray*}
where $\delta_0 := 2\exp\pth{-\frac{c_1\e^2k_2\rho(A^{*})}{(1-p)^2}}$.
\end{cor}

The next two results analyze the \emph{Stress} Metric, $\Lambda_S$.\cite{Bartal1} proved the following bound on \emph{Stress} if pure Random Map is applied:
\begin{thm}[Bartal et al.~\cite{Bartal1}]\label{bartal} Let $P\subset \R^D$ be a finite point set, $q\geq 2$, and $G:\R^D\to \R^{d}$ be a Gaussian random map. Then with probability at least $1/2$, the $q$-norm stress of the point set $P$ under the map $G$ satisfies 
        \begin{eqnarray*}
            \Lambda^{(q)}_S(P) &\leq &2\sqrt{\frac{3}{e}+\frac{3e^2}{2}}\sqrt{q/d} \\ & \leq &6.2\sqrt{q/d} \;\;= \;\;O(\sqrt{q/d}).
        \end{eqnarray*}

                    In particular, the $2$-norm stress, $\Lambda_S(P)$, satisfies $\Lambda_S(P) = O(\sqrt{1/d}).$
         \end{thm}

\begin{lem}\label{data-difference}
Given points $\mathbf{x}_1,\mathbf{x}_2,\dots, \mathbf{x}_n \in \mathbb{R}^{D}$ and data matrix $A$. Let $\mathbf{d}_{ij}=\mathbf{x}_i-\mathbf{x}_j$ , then:
$$\sum_{j<i}^n ||\mathbf{d}_{ij}||^2=n\sum_{j<i}^n||\mathbf{x}_i||^2 = n ||A||_{F}^{2}$$
\end{lem}
\begin{thm}[\emph{Stress} Bound]\label{stressBound}
Given a set of points $\mathbf{x}_1,\dots,\mathbf{x}_n$, $k_1$ and $k_2$, let application of the \textit{DiffRed} algorithm return the points $\tilde{\mathbf{x}_1},\dots,\tilde{\mathbf{x}_n}$. Then with probability at least $1/2$,  
$$\Lambda_S=O\left(\sqrt{\frac{1-p}{k_2}}\right)$$
\end{thm}
\begin{prf}
By definition, the value of \emph{Stress} is: 
    $$\Lambda_S^2=\frac{\sum_{i,j}(||\mathbf{d}_{ij}||-||\mathbf{\tilde{d}}_{ij}||)^2}{\sum||\mathbf{d}_{ij}||^2}.$$
    Now, since $\mathbf{d}_{ij}=\mathbf{x}_i-\mathbf{x}_j$ and $\mathbf{\Tilde{d}}_{ij}=\mathbf{\Tilde{x}}_i-\mathbf{\Tilde{x}}_j$, i.e.,  we can break them into two components that are \textbf{orthogonal} to each other:
    $$\mathbf{d}_{ij}=\mathbf{d}_{ij}^{(Z)}+\mathbf{d}_{ij}^{(Y)} \text{    and    }
    \mathbf{\tilde{d}}_{ij}=\mathbf{\tilde{d}}_{ij}^{(Z)}+ \mathbf{\tilde{d}}_{ij}^{(R)}.$$
  
    Here $\mathbf{d}_{ij}^{(Z)}, \mathbf{\tilde{d}}_{ij}^{(Z)} \in S_{k_1}$, the best-fit Subspace of rank $k_1$ and $\mathbf{d}_{ij}^{(Y)},\mathbf{\tilde{d}}_{ij}^{(R)} \in \mathbb{R}^d\setminus S_{k_1}$, the residual space. By using first $k_1$ principal components, \textit{DiffRed} ensures that $\mathbf{d}_{ij}^{(Z)} = \mathbf{\tilde{d}}_{ij}^{(Z)}$. It follows that    
    \begin{eqnarray*}
        ||\mathbf{d}_{ij}||-||\mathbf{\tilde{d}}_{ij}||& =\\  \sqrt{ ||\mathbf{d}_{ij}^{(Z)}||^2+||\mathbf{d}_{ij}^{(Y)}||^2} - \sqrt{||\mathbf{d}_{ij}^{(Z)}||^2 +||\mathbf{\tilde{d}}_{ij}^{(R)}||)^2}.&
    \end{eqnarray*}
    
    In supplementary material we prove the following useful inequality:
    \begin{equation}\label{stress-ineq}
        (\sqrt{a^2+b^2}-\sqrt{a^2+c^2})^2\leq (b-c)^2
    \end{equation}
Plugging $a=||\mathbf{d}_{ij}^{(Z)}||=||\mathbf{\tilde{d}}_{ij}^{(Z)}||$, $b=||\mathbf{d}_{ij}^{(Y)}||$ and $c=||\mathbf{\tilde{d}}_{ij}^{(R)}||$ helps us obtain the following bound on \emph{Stress}:
     \begin{equation}\label{pre-final}
        \Lambda_S^2\leq \frac{\sum_{i,j}(||\mathbf{d}_{ij}^{(Y)}||-||\mathbf{\tilde{d}}_{ij}^{(R)}||)^2}{\sum_{i,j}||\mathbf{d}_{ij}||^2}
    \end{equation}
Using Lemma \ref{data-difference},
\begin{eqnarray*}
    \sum_{i,j}||\mathbf{d}_{ij}||^2=n ||A||_F^2 \text{ and }\\  \sum_{i,j}||\mathbf{d}_{ij}^{(Y)}||^2=n ||A^{*}||_F^2 = (1-p)||A||_F^2  
\end{eqnarray*}
because $A^{*}$ is the residual matrix. Now, from  and these relations, it follows that:
    $$\frac{\sum_{i,j}||\mathbf{d}_{ij}||^2}{\sum_{i,j}||\mathbf{d}_{ij}^{(Y)}||^2}=\frac{1}{1-p}$$
    Using this in equation \ref{pre-final} we get:
    $$\Lambda_S^2\leq (1-p)\left(\frac{\sum_{i,j}(||\mathbf{d}_{ij}^{(Y)}||-||\mathbf{\tilde{d}}_{ij}^{(R)}||)^2}{\sum_{i,j}||\mathbf{d}_{ij}^{(Y)}||^2}\right)$$
    The RHS is simply now ($1-p$) times $\Lambda_S^2(R)$ which is the \emph{Stress} between the residual matrix $A-A_{k_1}$ and the matrix $R$. Now the statement of the Theorem follows from 
    Theorem \ref{bartal}. 
    $\qed$
\end{prf}

\begin{cor}[\emph{Stress} Bound, Monte Carlo version]\label{cor:stressBound}
Given a set of points $\mathbf{x}_1,\dots,\mathbf{x}_n$, $k_1$ and $k_2$, let application of the \textit{DiffRed} algorithm return the points $\tilde{\mathbf{x}_1},\dots,\tilde{\mathbf{x}_n}$, and given $\eta>0$, then the probability that in $\eta$ Monte Carlo iterations, 
the \emph{Stress} exceeds $O\pth{\frac{1-p}{k_2}}$, is at most  
$$\Prob{\Lambda_S \geq O\left(\sqrt{\frac{1-p}{k_2}}\right)}\leq \exp\pth{-\eta \ln 2}.$$
\end{cor}

\paragraph{Complexity Analysis}
   The complexity of the \textit{DiffRed} algorithm \ref{DiffRed-algo} is $O(Dn\cdot\min\{D,n\}+\eta nk_2D)$ which suggests that $\eta$ can be chosen of the order of $\frac{\min\{D,n\}}{k_2}$ to avoid adding more complexity than what is needed for $k_1$-rank approximation. (ref. Supplementary Material Section \ref{sec:complexity})

\section{Experiments}
We have extensively evaluated \textit{DiffRed} on various real-world datasets for stress and \emph{M1} distortion metrics\footnote{Code: \href{https://github.com/S3-Lab-IIT/DiffRed}{\texttt{\textcolor{blue}{https://github.com/S3-Lab-IIT/DiffRed}}}}. We first discuss the datasets, followed by the experimental setup, results, and various inferences. 

\subsection{Datasets}

\begin{table}[H]
\centering
\captionsetup{font=small}
\resizebox{\columnwidth}{!}{%
\begin{tabular}{|l|l|l|l|l|l|}
\hline
    \textbf{Name} & $\mathbf{D}$ & $\mathbf{n}$ & \textbf{Type} & $\mathbf{\rho}$ & \textbf{Domain} \\ \hline
    Bank & 17 & 45211 & Low & 1.48 & Finance \\ \hline
    Hatespeech & 100 & 3221 & Low & 11.00 & NLP \\ \hline
    F-MNIST & 784 & 60000 & Low & 2.68 & Image \\ \hline
    Cifar10  & 3072 & 50000 & Medium & 6.13 & Image \\ \hline
    geneRNASeq & 20.53K & 801 & Medium & 1.12 & Biology \\ \hline
    Reuters30k & 30.92K & 10788 & Medium & 14.50 & NLP \\ \hline
    APTOS 2019 & 509k & 13000  & High & 1.32 & Healthcare \\ \hline
    DIV2K & 6.6M & 800 &  Very High & 8.39 & High Res Image \\ \hline
\end{tabular}
}

\caption{Summary of the datasets used with their respective type based on dimensionality}
\label{tab:datasets}
\end{table}
\squeezeup
\vspace*{-\baselineskip}
Table \ref{tab:datasets} summarizes the datasets used for our experiments. Our datasets span a wide range of dimensionality,  application domains and stable ranks. Bank \cite{misc_bank_marketing_222} is a binary classification dataset of the marketing campaign of a Portuguese banking institution. Fashion MNIST [F-MNIST] \cite{xiao2017fashionmnist} is a multiclass classification dataset of grayscale images of 10 different kinds of fashion products. Cifar10 \cite{Krizhevsky2009LearningML} is a dataset of RGB images of various objects. geneRNASeq \cite{misc_gene_expression_cancer_rna-seq_401} is a random extraction of gene expressions of patients having five different types of tumors. Reuters30k is the TF-IDF representation of the Reuters-21578 dataset \cite{misc_reuters-21578_text_categorization_collection_137} which is a collection of documents consisting of financial news articles that appeared on Reuters newswire in 1987. APTOS 2019 \cite{aptos2019-blindness-detection} is a dataset of retina images used for predicting the severity of diabetic retinopathy. DIV2k \cite{8014884,Agustsson_2017_CVPR_Workshops} is a collection of high-resolution 2K images. 

\subsection{Experimental Setup}
For experimentation, we used a workstation with Intel(R) Xeon(R) Gold 5218 CPU @ 2.30GHz, with NVIDIA RTX A6000 GPU, and a shared commodity cluster. We used Python and slurm-based shell scripting to run our experiments. To speed up our experiments (especially computation of \emph{Stress}), our entire codebase was written to leverage multiprocessing.\\
\textbf{\textit{Pre-processing:}} We scale each dataset to zero mean and normalize the examples to be vectors of unit norm. We convert the datasets into their vector representations using various standard techniques like label encoding, tf-idf, etc. wherever applicable. \\
\textbf{\textit{Computing Embeddings:}} \prarabdh{We have compared \textit{DiffRed} to various dimensionality reduction algorithms. Our choice of algorithms was based on the results presented by \cite{Espadoto2019TowardAQ}}. For each of these techniques, we have tried to use the most appropriate implementation wherever possible. For T-SNE, we used T-SNE CUDA \cite{chan2019gpu}: a GPU version of T-SNE to compute the embeddings. For UMap, we used the official UMap implementation. For KernelPCA, SparsePCA and PCA we used \textit{scikit-learn's} \cite{scikit-learn} implementation. For RMap, we have used our own implementation. We have used $\alpha=20$ Gaussian RMaps on each target dimension with the same multiplicative factor and hyperparameters as specified in the \textit{DiffRed} algorithm [\ref{DiffRed-algo}]. We have used $\alpha=20$ random maps to build and report our 95\% confidence interval. For generating random Gaussian vectors and for our own \textit{DiffRed}, we have mainly relied on numPy's routines. \prarabdh{In our experiments, we have done hyperparameter tuning using a grid search to justify our theory and show various observations. We have taken the best hyperparameters reported by \cite{Espadoto2019TowardAQ} as the starting point for the grid search. Finally, after hyperparameter tuning, we compare \textit{DiffRed} to the best \emph{Stress} found for each target dimension for each Dimensionality Reduction technique.}

\subsection{Experiment Results}\label{sec:exp-res}

\prarabdh{To evaluate the performance of \textit{DiffRed}, we compute the \emph{Stress} and \emph{M1} distortion metrics on different datasets for different target dimensions. As a part of our experiments, we perform a grid search on different values of $k_1$ and $k_2$. We use the results of the grid search to justify our method of choosing $k_1$ and $k_2$ for a given target dimension (discussed in Section \ref{sec:Discusssion}) and to validate our theory (discussed in Sections \ref{sec:M1} and \ref{sec:stress}).}
\squeezeup
\subsubsection{Insights on \emph{M1}}\label{sec:M1}
\prarabdh{In this section, we discuss RMap and \textit{DiffRed} in light of Corollary \ref{cor:RMapM1Bound} and \ref{cor:M1Bound} respectively. The main observations are as follows:} 
\squeezeup
\input{tables/m1-results}

\prarabdh{ \paragraph{Observation 1 \label{obs:m1-a}} In Figure \ref{fig:m1-plot-rmap1}, we see that for a fixed target dimension of 10, datasets\footnote{except Bank, which has a low dimensionality to begin with.} with higher stable rank have lower $\Lambda_{M_1}$. In Figure \ref{fig:m1-plot-rmap2}, we see that for Reuters30k (i.e., fixed stable rank of $\rho=14.50$), higher target dimensions cause lesser \emph{M1} distortion. These empirical observations are in agreement with Corollary \ref{cor:RMapM1Bound}, where $\Prob{\Lambda_{M1}\geq \e}$ depends on the negative exponent of $d\times \rho(A)$. Therefore, for minimization of $\Lambda_{M_1}$ we require either a high stable rank or a high target dimension. Since stable rank incorporates the spread in data, we observe that, \textbf{contrary to the popular belief, Random Maps are not data agnostic.} }  
 
 \input{figure_files/m1-plot-rmap1}
 \input{figure_files/m1-plot-rmap2}
 \input{tables/stress-results}

\prarabdh{\paragraph{Observation 2 \label{obs:m1-b}} From Table \ref{tab:m1-results}, we observe that \textit{DiffRed} has the best values for \emph{M1} across all datasets. Detailed results on \emph{M1} distortion are deferred to the supplementary material.}

\prarabdh{ \paragraph{Observation 3 \label{obs:m1-c}} An interesting observation is that the \emph{M1} metric is insensitive to the choice of $k_1$ and $k_2$. To measure the sensitivity of $\Lambda_{M_1}$ on $k_1$ and $k_2$, we define the following quantity, $\beta$:
$$\beta=\underset{d \in \{10,20,30,40\}}{\proc{Average}}\left(\underset{k_1,k_2}{Var}(\Lambda_{M_1})\right)$$
$\beta$ is the average (over target dimensions $d$) of the variance observed in $\Lambda_{M_1}$ for different pairs of $k_1$ and $k_2$. In essence, $\beta$ is a measure of the sensitivity of $\Lambda_{M_1}$ w.r.t. $k_1$ and $k_2$ for a given dataset. We evaluated $\beta$ for different values of $k_1$ and $k_2$, observed that the average of $\beta$ across all datasets, $\langle \beta \rangle \approx 1.54 \times 10^{-6}$ (ref. Table 1 in the supplementary material).\\
The low sensitivity to $k_1$ and $k_2$ follows from Corollary \ref{cor:M1Bound}, where for a constant $\eta$, $\Prob{\Lambda_{M1}\geq \e}$ depends on the negative exponent of $\frac{k_2\rho(A^{*})}{(1-p)^2}$. We can consider two cases now: (i) $k_2$ is high and (ii) $k_2$ is low. As illustrated by Figure \ref{fig:m1-product-plot-diffred}, the exponent term remains sufficiently high for different stable ranks if $k_2$ is high (i.e., Case (i) holds). Now, for the second case, we make another observation from Figure \ref{fig:stable-rank-plots} that stable rank increases with $k_1$. Since a low $k_2$ value implies a high $k_1$ value ($k_1=d-k_2$), the exponent term remains high because of the high stable rank. }

 \input{figure_files/m1-product-plot-diffred}

\squeezeup
\vspace*{-\baselineskip}

\subsubsection{Insights on \emph{Stress}} \label{sec:stress}
\prarabdh{In this section, we discuss the various observations we make in context of the \emph{Stress} metric. The major observations are as follows:}

\prarabdh{\paragraph{Observation 1 \label{obs:stress-a}} We observe that \textit{DiffRed} achieves the best values of \emph{Stress} among all other algorithms (Table \ref{tab:stress-results}). We evaluated the \emph{Stress} metric on all our datasets for target dimensions 10 to 40 for different values of $k_1$ and $k_2$. In Table \ref{tab:stress-results} we compare the empirically obtained best values with the best values for other commonly used Dimensionality Reduction techniques. But we observe that a grid search to determine optimal $k_1$ and $k_2$ for a given target dimension and dataset is not required, as we will discuss in Section \ref{sec:opt-hyp} (Choice of hyperparameters).}
\\
\prarabdh{We note that among all datasets, \textit{DiffRed} consistently achieves the lowest \emph{Stress} values, even when the dimensionality is very high. Sparse-PCA remains close to PCA while Kernel-PCA has a higher \emph{Stress} value. Techniques such as UMap (manifold approximation) and T-SNE (which preserves neighborhoods) do not perform well on distance based metrics. Therefore, to be fair to them, we have included versions UMap2 and T-SNE2 in Table \ref{tab:stress-results}. These versions are \textit{energy-matched} with the original data, i.e., they have been re-scaled such that their Frobenius norm (energy) matches that of the original data (i.e., $\Lambda_{M_1}=0$).}
\input{figure_files/stress-plot-rmap}
\squeezeup
\vspace*{-\baselineskip}
\prarabdh{\paragraph{Observation 2\label{obs:stress-b}} In accordance with Theorem \ref{bartal}, Random Maps preserve \emph{Stress} better if more target dimensions are allowed. In Figure \ref{fig:stress-plot-rmap}, we see that RMap benefits in context of the \emph{Stress} metric if more target dimensions are allowed. This is the behavior one would expect from Theorem \ref{bartal}, which bounds \emph{Stress} of random map as $O\left(\sqrt{\frac{1}{d}}\right)$.}

. 
\squeezeup
\vspace*{-\baselineskip}

\paragraph{\prarabdh{Observation 3\label{obs:stress-c}}} \prarabdh{From Figure \ref{fig:stress-plot-pca}, we make two observations: i. PCA benefits in context of \emph{Stress} if more target dimensions are allowed and ii. The general trend of PCA is to perform better for datasets whose stable rank is low to begin with. (see Table \ref{tab:stress-results}). Complementary to this observation, we also note from Table \ref{tab:stress-results} that the general trend for RMap is to perform better for datasets that have a high stable rank to begin with.}

\input{figure_files/stress-plots-pca}
\squeezeup
\vspace*{-\baselineskip}
\paragraph{\prarabdh{Observation 4 \label {obs:stress-d}}} \prarabdh{From Figure \ref{fig:stress-plot-diffred}, we note that with \textit{DiffRed}, we see an improvement in preserving \emph{Stress} as more target dimensions are allowed (as suggested by Theorem \ref{stressBound}, $O\left(\sqrt{\frac{1-p}{k_2}}\right)$). However, in our grid search experiments on \emph{Stress} (as described in Observation \ref{obs:stress-a} above), we observe that, unlike \emph{M1}, \emph{Stress} is, in fact, sensitive to the choice of $k_1$ and $k_2$ (For discussion on the choice of these hyperparameters, ref. Section \ref{sec:opt-hyp} [Choice of hyperparameters] ). In conclusion, if a particular downstream tasks benefits from lower \emph{Stress}, one may simply allow more target dimensions.}

\input{figure_files/stress-plots-diffred}
\vspace*{-\baselineskip}

\subsection{Discussion} \label{sec:Discusssion}
In this section, first we validate our theory results from our experiments and then we discuss the choice of hyperaparameters and possible applications of \textit{DiffRed}

\textit{DiffRed} generally performs better than RMap because of the additional $\sqrt{1-p}$ factor in the \emph{Stress} bound [Theorem \ref{stressBound}]. 
Additionally, Figure \ref{fig:bound-exp-plot} below shows that both our bounds [Theorem \ref{M1Bound} and Theorem \ref{stressBound}] hold good in our experiments. 
\input{figure_files/bound-exp-plot}
\squeezeup
\vspace*{-\baselineskip}
\paragraph{Choice of hyperparameters} \label{sec:opt-hyp}
As described in Section \ref{sec:M1}, the \emph{M1} metric is not sensitive to values of $k_1$ and $k_2$, therefore, most values of $k_1$ and $k_2$ minimize the \emph{M1} distortion. As for \emph{Stress}, we have observed in our grid search experiments \prarabdh{[Figure \ref{fig:theory-opt-vs-exp-opt}]} that values of $k_1$ and $k_2$ that minimize the theoretical bound usually give \emph{Stress} values that are close to the empirically observed minima.  \prarabdh{Therefore, one may simply choose the $k_1$, $k_2$ pair that minimizes the theoretical bound (i.e., the value of $\sqrt{\frac{1-p}{k_2}}$)  for minimizing \emph{Stress}. Computing the bound value is inexpensive, and therefore, one may simply iterate over all combinations of $k_1$ and $k_2$ for a given target dimension to find the minima. }

\input{figure_files/theory-opt-vs-exp-opt}

\input{figure_files/stable-rank-plots}
\squeezeup
\vspace*{-\baselineskip}
\paragraph{Effect of Monte Carlo iterations}
In our experiments (ref. Supplementary Material) we found that by increasing $\eta$, we can find Random Maps that further reduce \emph{M1} metric. It is very interesting to note that such Random Maps achieve lower \emph{Stress}. This justifies the selection of Random Maps based on minimization of \emph{M1} in the \textit{DiffRed} algorithm.
\squeezeup

\section{Conclusion}

In this paper, we design a new dimensionality reduction algorithm, \textit{DiffRed} and obtain new bounds for \emph{M1} and \emph{Stress} metrics that are tighter than currently known results for Random Maps. \textit{DiffRed} uses the notion of stable rank in choosing the directions for projecting the dataset. When the stable rank of a dataset is high to begin with, it emphasizes random maps. When the stable rank of the dataset is low to begin with, it first chooses enough number of principal components so that the stable rank of the residual matrix increases and then uses random maps. \prarabdh{Therefore, by incorporating stable rank (structure of data) into our bound, we have shown how dimensionality reduction can be guided by stable rank, thereby reducing the required target dimension.} Through extensive experiments on real-world datasets, we have shown that \textit{DiffRed} obtains significant reduction in \emph{M1} and \emph{Stress} as compared to well known dimensionality reduction algorithms.
As a part of future work, researchers can explore the effectiveness of \textit{DiffRed} to various applications such as Clustering, Visualization, Nearest Neighbor Search, etc., where high dimensionality often becomes a bottleneck and a global-structure-preserving representation is required in lower dimensions. 

\paragraph{Acknowledgement} Kunal Dutta's work on this project was supported by the NCN SONATA grant nr. 2019/35/D/ST6/04525.
\bibliography{ref}
 \begin{enumerate}

\section{Checklist}
 \item For all models and algorithms presented, check if you include:
 \begin{enumerate}
   \item A clear description of the mathematical setting, assumptions, algorithm, and/or model. [Yes]
   \item An analysis of the properties and complexity (time, space, sample size) of any algorithm. [Yes]
   \item (Optional) Anonymized source code, with specification of all dependencies, including external libraries. [Yes]
 \end{enumerate}

 \item For any theoretical claim, check if you include:
 \begin{enumerate}
   \item Statements of the full set of assumptions of all theoretical results. [Yes]
   \item Complete proofs of all theoretical results. [Yes]
   \item Clear explanations of any assumptions. [Yes]     
 \end{enumerate}

 \item For all figures and tables that present empirical results, check if you include:
 \begin{enumerate}
   \item The code, data, and instructions needed to reproduce the main experimental results (either in the supplemental material or as a URL). [Yes]
   \item All the training details (e.g., data splits, hyperparameters, how they were chosen). [Yes]
         \item A clear definition of the specific measure or statistics and error bars (e.g., with respect to the random seed after running experiments multiple times). [Yes]
         \item A description of the computing infrastructure used. (e.g., type of GPUs, internal cluster, or cloud provider). [Yes]
 \end{enumerate}

 \item If you are using existing assets (e.g., code, data, models) or curating/releasing new assets, check if you include:
 \begin{enumerate}
   \item Citations of the creator If your work uses existing assets. [Yes]
   \item The license information of the assets, if applicable. [Yes]
   \item New assets either in the supplemental material or as a URL, if applicable. [Yes]
   \item Information about consent from data providers/curators. [Not Applicable]
   \item Discussion of sensible content if applicable, e.g., personally identifiable information or offensive content. [Not Applicable]
 \end{enumerate}

 \item If you used crowdsourcing or conducted research with human subjects, check if you include:
 \begin{enumerate}
   \item The full text of instructions given to participants and screenshots. [Not Applicable]
   \item Descriptions of potential participant risks, with links to Institutional Review Board (IRB) approvals if applicable. [Not Applicable]
   \item The estimated hourly wage paid to participants and the total amount spent on participant compensation. [Not Applicable]
 \end{enumerate}

 \end{enumerate}
\input{MainPaper/supplement}
\end{document}

%% file: figure_files/stable-rank-intuition.tex
\begin{figure}[h]
\captionsetup{font=small}
    \centering
    \includegraphics[scale=0.25]{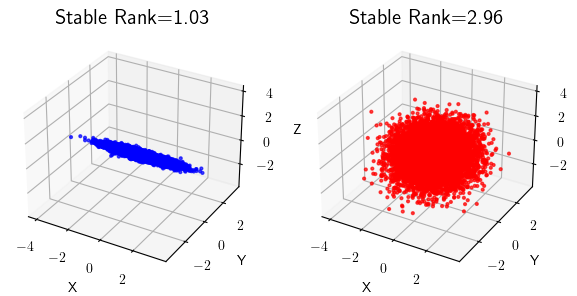}
    \caption{Stable rank as a measure of "spread" in data in a 3-D example.}
    \label{fig:stable-rank-intuition}
\end{figure}

%% file: tables/m1-results.tex
\begin{table*}[t]
\centering
\captionsetup{font=small}

\begin{tabular}{|l|c|c|ccccccc|}

\hline
\multirow{2}{*}{\textbf{Dataset}} &
  \multirow{2}{*}{$D$} &
  \multirow{2}{*}{$d$} &
  \multicolumn{7}{c|}{$\Lambda_{M_1}$} \\ \cline{4-10} 
 &
   &
   &
  \multicolumn{1}{c|}{DiffRed} &
  \multicolumn{1}{c|}{PCA} &
  \multicolumn{1}{c|}{RMap} &
  \multicolumn{1}{c|}{S-PCA} &
  \multicolumn{1}{c|}{K-PCA} &
  \multicolumn{1}{c|}{UMap} &
  \begin{tabular}[c]{@{}c@{}}T-SNE\\ ($d=2$)\end{tabular} \\ \hline
Bank &
  17 &
  5 &
  \multicolumn{1}{c|}{\textbf{2.82e-05}} &
  \multicolumn{1}{c|}{0.54} &
  \multicolumn{1}{c|}{0.38} &
  \multicolumn{1}{c|}{0.58} &
  \multicolumn{1}{c|}{0.95} &
  \multicolumn{1}{c|}{94.89} &
  2659.70 \\ \hline
Hatespeech &
  100 &
  10&
  \multicolumn{1}{c|}{\textbf{1.91e-04}} &
  \multicolumn{1}{c|}{0.66} &
  \multicolumn{1}{c|}{0.06} &
  \multicolumn{1}{c|}{0.68} &
  \multicolumn{1}{c|}{0.99} &
  \multicolumn{1}{c|}{240.50} &
  2298.09\\ \hline
FMnist &
  784 &
  10 &
  \multicolumn{1}{c|}{\textbf{1.92e-04}} &
  \multicolumn{1}{c|}{0.60} &
  \multicolumn{1}{c|}{0.11} &
  \multicolumn{1}{c|}{0.64} &
  \multicolumn{1}{c|}{1.00} &
  \multicolumn{1}{c|}{241.35} &
  829.54 \\ \hline
Cifar10 &
  3072 &
  10 &
  \multicolumn{1}{c|}{\textbf{1.31e-04}} &
  \multicolumn{1}{c|}{0.49} &
  \multicolumn{1}{c|}{0.09} &
  \multicolumn{1}{c|}{0.54} &
  \multicolumn{1}{c|}{1.00} &
  \multicolumn{1}{c|}{166.84} &
  604.71 \\ \hline
geneRNASeq &
  20.5k &
  10 &
  \multicolumn{1}{c|}{\textbf{7.96e-05}} &
  \multicolumn{1}{c|}{0.94} &
  \multicolumn{1}{c|}{0.31} &
  \multicolumn{1}{c|}{0.95} &
  \multicolumn{1}{c|}{1.00} &
  \multicolumn{1}{c|}{328.72} &
  8,761.41 \\ \hline
Reuters30k &
  30.9k &
  10 &
  \multicolumn{1}{c|}{\textbf{1.27e-04}} &
  \multicolumn{1}{c|}{0.88} &
  \multicolumn{1}{c|}{0.03} &
  \multicolumn{1}{c|}{0.88} &
  \multicolumn{1}{c|}{1.00} &
  \multicolumn{1}{c|}{196.97} &
  2393.31 \\ \hline
APTOS 2019 &
  509k &
   10&
  \multicolumn{1}{c|}{\textbf{4.09e-05}} &
  \multicolumn{1}{c|}{0.81} &
  \multicolumn{1}{c|}{0.24} &
  \multicolumn{1}{c|}{-} &
  \multicolumn{1}{c|}{-} &
  \multicolumn{1}{c|}{-} &
  - \\ \hline
DIV2k &
  6.6M &
  10 &
  \multicolumn{1}{c|}{\textbf{7.07e-05}} &
  \multicolumn{1}{c|}{0.66} &
  \multicolumn{1}{c|}{0.05} &
  \multicolumn{1}{c|}{-} &
  \multicolumn{1}{c|}{-} &
  \multicolumn{1}{c|}{-} &
  - \\ \hline
\end{tabular}
\caption{Comparison of the \emph{M1} metric.
Note that $k_1=2$ and $k_2=3$ for Bank and  $k_1=6$ and $k_2=4$ for other datasets. \prarabdh{For APTOS and DIV2k, \emph{M1} is evaluated for only PCA, RMap, and \textit{DiffRed} due to memory limitations.}}
\label{tab:m1-results}
\end{table*}

%% file: figure_files/m1-plot-rmap1.tex
\begin{figure}[h]
\captionsetup{font=small}
    \centering
   \includegraphics[width=0.45\textwidth]{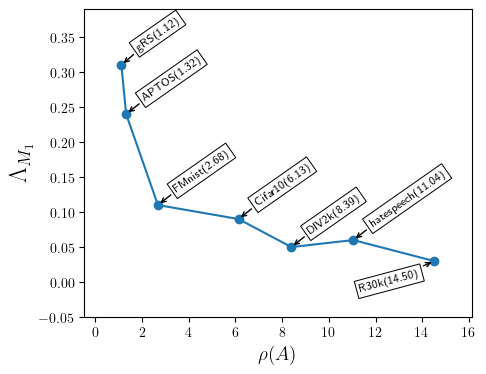}
    \caption{Dependence of \emph{M1} for Random Maps on stable rank described in Corollary \ref{cor:RMapM1Bound}.($d=10$) [\textit{gRS}: geneRNASeq, \textit{R30k}: Reuters30k]}
    \label{fig:m1-plot-rmap1}
\end{figure}

%% file: figure_files/m1-plot-rmap2.tex
\begin{figure}[h]
\captionsetup{font=small}
    \centering
   \includegraphics[width=0.25\textwidth]{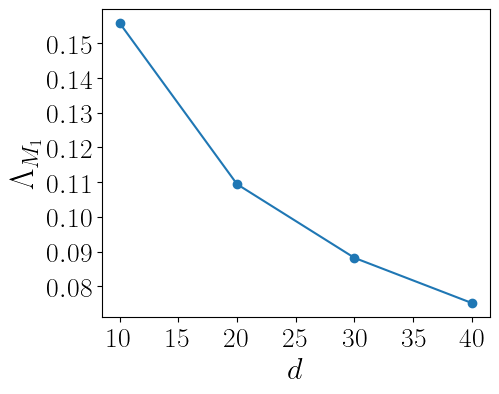}
    \caption{Variation of \emph{M1} with target dimension for Reuters30k}
    \label{fig:m1-plot-rmap2}
\end{figure}

%% file: tables/stress-results.tex
\begin{table*}[t]
\centering
\captionsetup{font=small}
\resizebox{1.9\columnwidth}{!}{%
\begin{tabular}{|l|l|c|ccccccccc|}
\hline
 &
   &
   &
  \multicolumn{9}{c|}{$\Lambda_S$} \\ \cline{4-12} 
\multirow{-2}{*}{\textbf{Dataset}} &
  \multirow{-2}{*}{$D$} &
  \multirow{-2}{*}{$d$} &
  \multicolumn{1}{c|}{DiffRed} &
  \multicolumn{1}{c|}{PCA} &
  \multicolumn{1}{c|}{RMap} &
  \multicolumn{1}{c|}{S-PCA} &
  \multicolumn{1}{c|}{K-PCA} &
  \multicolumn{1}{c|}{UMap} &
  \multicolumn{1}{c|}{UMap2} &
  \multicolumn{1}{c|}{\begin{tabular}[c]{@{}c@{}}T-SNE\\ ($d=2$)\end{tabular}} &
  \begin{tabular}[c]{@{}c@{}}T-SNE2\\ ($d=2$)\end{tabular} \\ \hline
Bank &
  17 &
  6 &
  \multicolumn{1}{c|}{\textbf{0.02}} &
  \multicolumn{1}{c|}{0.03} &
  \multicolumn{1}{c|}{0.17} &
  \multicolumn{1}{c|}{0.04} &
  \multicolumn{1}{c|}{0.47} &
  \multicolumn{1}{c|}{7.07} &
  \multicolumn{1}{c|}{0.35} &
  \multicolumn{1}{c|}{52.44} &
  0.72 \\ \hline
Hatespeech &
100 &
10 &
 \multicolumn{1}{c|}{\textbf{0.15}} &
  \multicolumn{1}{c|}{0.36} &
  \multicolumn{1}{c|}{0.16} &
  \multicolumn{1}{c|}{0.36} &
  \multicolumn{1}{c|}{0.65} &
  \multicolumn{1}{c|}{5.29} &
  \multicolumn{1}{c|}{0.46} &
  \multicolumn{1}{c|}{32.86} &
  0.38 \\ \hline
FMnist &
  784 &
  10 &
  \multicolumn{1}{c|}{\textbf{0.12}} &
  \multicolumn{1}{c|}{0.19} &
  \multicolumn{1}{c|}{0.15} &
  \multicolumn{1}{c|}{0.21} &
  \multicolumn{1}{c|}{0.68} &
  \multicolumn{1}{c|}{4.02} &
  \multicolumn{1}{c|}{0.42} &
  \multicolumn{1}{c|}{24.49}&
  0.38 \\ \hline
Cifar10 &
  3072 &
  10 &
  \multicolumn{1}{c|}{\textbf{0.13}} &
  \multicolumn{1}{c|}{0.21} &
  \multicolumn{1}{c|}{0.16} &
  \multicolumn{1}{c|}{0.24} &
  \multicolumn{1}{c|}{0.69} &
   \multicolumn{1}{c|}{1.26} &
   \multicolumn{1}{c|}{0.60} &
   \multicolumn{1}{c|}{16.88} &
  0.31 \\ \hline
geneRNASeq &
  20.5k &
  10 &
  \multicolumn{1}{c|}{\textbf{0.13}} &
  \multicolumn{1}{c|}{0.21} &
  \multicolumn{1}{c|}{0.16} &
  \multicolumn{1}{c|}{0.25} &
  \multicolumn{1}{c|}{0.70} &
  \multicolumn{1}{c|}{18.72} &
  \multicolumn{1}{c|}{0.47} &
  \multicolumn{1}{c|}{164.89} &
  1.21 \\ \hline
Reuters30k &
  30.9k &
  10 &
  \multicolumn{1}{c|}{\textbf{0.155}} &
  \multicolumn{1}{c|}{0.49} &
  \multicolumn{1}{c|}{0.157} &
  \multicolumn{1}{c|}{0.49} &
  \multicolumn{1}{c|}{0.71} &
   \multicolumn{1}{c|}{3.35} &
   \multicolumn{1}{c|}{0.44} &
   \multicolumn{1}{c|}{18.02} &
  0.31 \\ \hline
\multicolumn{1}{|l|}{APTOS 2019} &
  509k &
  \multicolumn{1}{l|}{10} &
  \multicolumn{1}{c|}{\textbf{0.10}} &
  \multicolumn{1}{c|}{0.12} &
  \multicolumn{1}{c|}{0.16} &
  \multicolumn{1}{c|}{-} &
  \multicolumn{1}{c|}{-} &
   \multicolumn{1}{c|}{-} &
   \multicolumn{1}{c|}{-} &
   \multicolumn{1}{c|}{-} &
  - \\ \hline
\multicolumn{1}{|l|}{DIV2k} &
  6.6M &
  \multicolumn{1}{l|}{10} &
  \multicolumn{1}{c|}{\textbf{0.14}} &
  \multicolumn{1}{c|}{0.31} &
  \multicolumn{1}{c|}{0.16} &
  \multicolumn{1}{c|}{-} &
  \multicolumn{1}{c|}{-} &
   \multicolumn{1}{c|}{-} &
   \multicolumn{1}{c|}{-} &
   \multicolumn{1}{c|}{-} &
  - \\ \hline
\end{tabular}
}
\caption{Comparison of \textit{DiffRed} with other dimensionality reduction algorithms in context of \emph{Stress}. For \textit{DiffRed}, the best \emph{Stress} from grid search is reported. For APTOS and DIV2k, \emph{Stress} is evaluated for only PCA, RMap, and \textit{DiffRed} due to memory limitations.}

\label{tab:stress-results}
\end{table*}

%% file: figure_files/m1-product-plot-diffred.tex
\begin{figure}[H]
    \centering
    \captionsetup{font=small}
    \begin{subfigure}[t]{0.23\textwidth}
    \centering
    \includegraphics[width=\textwidth]{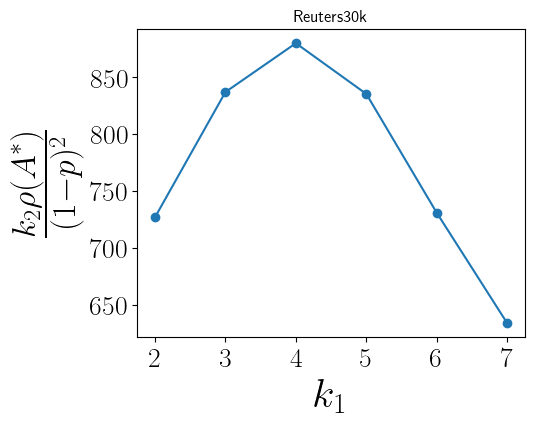}
    \caption{Reuters30k($\rho=14.50)$}
    \label{fig:m1-product-plot-diffred1}
\end{subfigure}%
~
\begin{subfigure}[t]{0.23\textwidth}
    \centering
    \includegraphics[width=\textwidth]{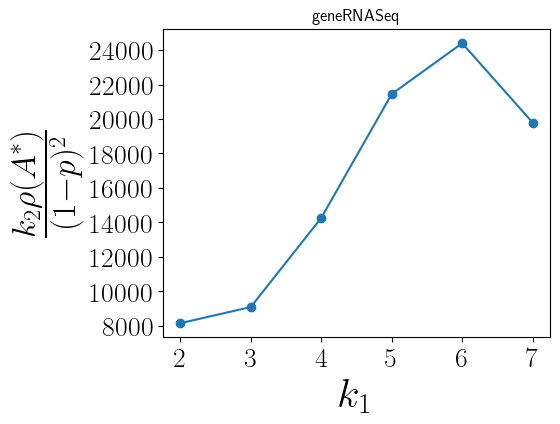}
    \caption{geneRNASeq ($\rho=1.12)$}
    \label{fig:m1-product-plot-diffred2}
\end{subfigure}
\caption{The exponent term $\frac{k_2\rho(A^{*})}{(1-p)^2}$ 
         remains high for different values of $k_1$. ($d=10$)}
         \label{fig:m1-product-plot-diffred}
\end{figure} 

%% file: figure_files/stress-plot-rmap.tex
\begin{figure}[H]
    \centering
    \captionsetup{font=small}
    \begin{subfigure}[t]{0.25\textwidth}
    \centering
    \includegraphics[width=\textwidth]{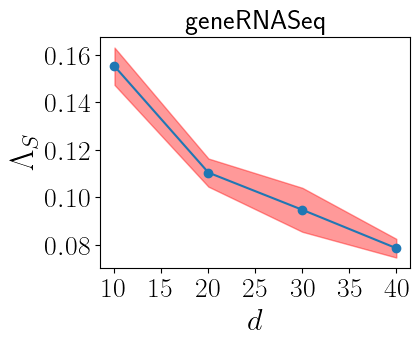}
    \caption{geneRNASeq($\rho=1.12$)}
    \label{fig:stress-plot-rmap1}
\end{subfigure}%
~
\begin{subfigure}[t]{0.25\textwidth}
    \centering
    \includegraphics[width=\textwidth]{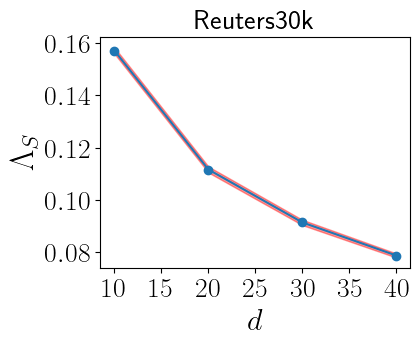}
    \caption{Reuters30k ($\rho=14.50$)}
    \label{fig:stress-plot-rmap2}
\end{subfigure}
\caption{ $\Lambda_S$ vs $d$ for RMap (95\% confidence interval in red).$\alpha=20$ RMaps were used to generate confidence interval. }
\label{fig:stress-plot-rmap}
\end{figure}

%% file: figure_files/stress-plots-pca.tex
\begin{figure}[H]
\captionsetup{font=small}
    \centering
    \includegraphics[width=0.30\textwidth]{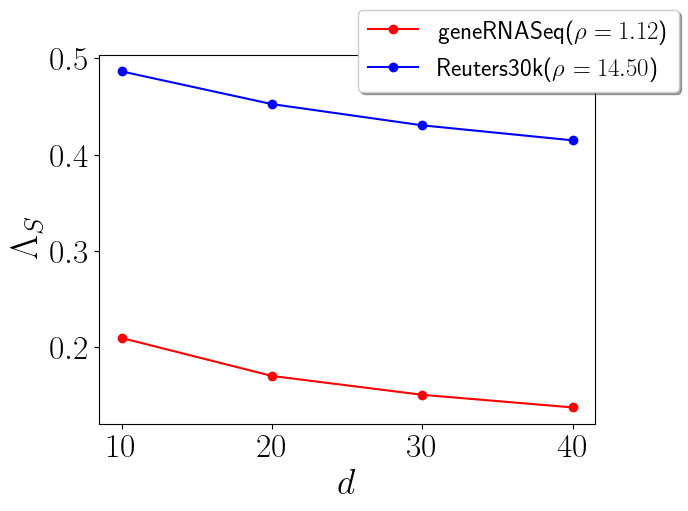}
    \caption{Plot showing how PCA benefits from more target dimensions ($d$) in context of \emph{Stress}.}
    \label{fig:stress-plot-pca}
\end{figure}

%% file: figure_files/stress-plots-diffred.tex
\begin{figure}[H]
    \centering
    \captionsetup{font=small}
    \begin{subfigure}[t]{0.25\textwidth}
    \centering
    \includegraphics[width=\textwidth]{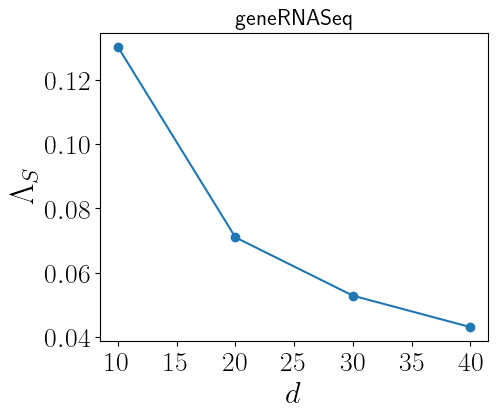}
    \caption{geneRNASeq($\rho=1.12$)}
    \label{fig:stress-plot-diffred1}
\end{subfigure}%
~
\begin{subfigure}[t]{0.25\textwidth}
    \centering
    \includegraphics[width=\textwidth]{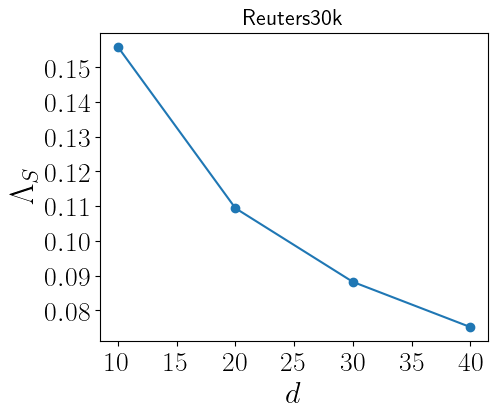}
    \caption{Reuters30k ($\rho=14.50$)}
    \label{fig:stress-plot-diffred2}
\end{subfigure}
\caption{$\Lambda_S$ vs $d$ when \textit{DiffRed} is used.}
\label{fig:stress-plot-diffred}
\end{figure}

%% file: figure_files/bound-exp-plot.tex
\begin{figure}[H]
    \centering
    \captionsetup{font=small}
    \begin{subfigure}[t]{0.25\textwidth}
    \centering
    \includegraphics[width=0.9\textwidth]{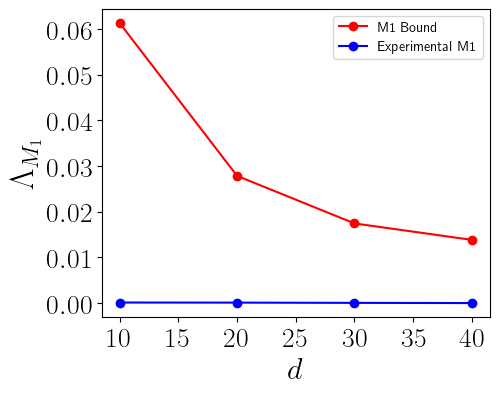}
    \caption{Comparison of the \emph{M1 Bound} [Theorem \ref{M1Bound}] and the experimentally observed \emph{M1}.}
    \label{fig:m1-bound-exp}
\end{subfigure}%
~
\begin{subfigure}[t]{0.25\textwidth}
    \centering
    \includegraphics[width=0.9\textwidth]{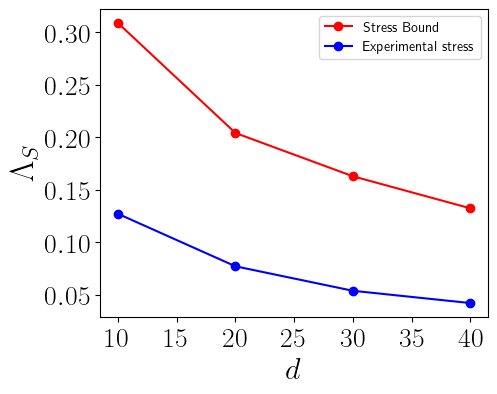}
    \caption{Comparison of the \emph{Stress bound} [Theorem \ref{stressBound}] and the experimentally observed \emph{Stress}.}
    \label{fig:stress-bound-exp}
\end{subfigure}
\caption{Comparison between theoretical bounds and empirical observations. }
\label{fig:bound-exp-plot}
\end{figure}

%% file: figure_files/theory-opt-vs-exp-opt.tex
\begin{figure}[H]
\captionsetup{font=small}
    \centering
    \includegraphics[scale=0.25]{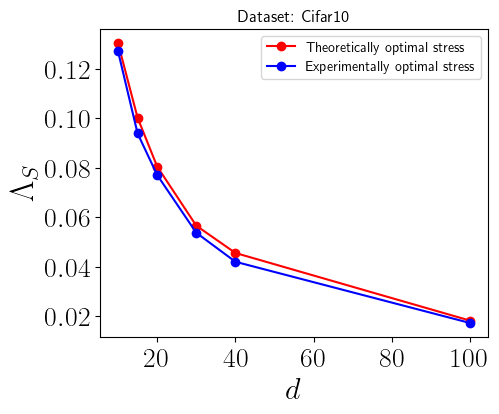}
    \caption{(For Cifar10) [Red]\emph{Stress} for $k_1$, $k_2$ values that minimize the bound in Theorem \ref{stressBound}. [Blue] \emph{Stress} at empirically optimal $k_1$ and $k_2$ values.  }
    \label{fig:theory-opt-vs-exp-opt}
\end{figure}

%% file: figure_files/stable-rank-plots.tex
\begin{figure}
    \centering
    \captionsetup{font=small}
    \includegraphics[scale=0.50]{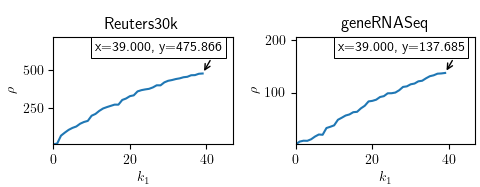}
    \caption{Plots of stable rank vs $k_1$. Plots for other datasets are in the supplementary material [Figure \ref{fig:sr-plots}].}
    \label{fig:stable-rank-plots}
\end{figure}

%% file: MainPaper/supplement.tex
\onecolumn
\aistatstitle{Supplementary Material}

\squeezeup
\vspace*{-\baselineskip}
\section{Proof of Theorem \ref{M1Bound}}
\begin{proof}

    From the \textit{DiffRed} algorithm presented in Section \ref{sec:dr-algo}, 
    $\Tilde{A}=[Z|R]$, so that $||\Tilde{A}||_F^2=||Z||_F^2+||R||_F^2$.
    Using the identity that $||Z||_F^2=\sum_{i=1}^{k_1} \sigma_i^2$ gives \\
    \begin{equation}
        \begin{split}
            \|\Tilde{A}\|_F^2 = &\sum_{i=1}^{k_1} \sigma_i^2+||R||_F^2 \;\;=\;\; p\sum_{i=1}^{r} \sigma_i^2+||R||_F^2 \\ = & p\|A\|_F^2+||R||_F^2. \label{1}
        \end{split}
    \end{equation}
    
Now, from Lemma \ref{l:rand-map-avg-dist}, we know that with probability at least $1-2\cdot\exp\pth{-c_1\e^2d\rho_A}$ we have:
$$|(\|R\|_F^2-\|A^{*}\|_F^2)|\leq \e\|A^{*}\|_F^2$$
Let $\frac{\e}{1-p}=\e'$. 
This implies,
\begin{equation}
   \Prob{(1-\e')\|A^{*}\|_F^2\leq \|R\|_F^2 \leq (1+\e')\|A^{*}\|_F^2} 
    \leq  2\cdot 
    \exp\pth{-\frac{c_1\e^2k_2\rho(A^{*})}{(1-p)^2}}
\end{equation}

Now, using this in equation \eqref{1}, we observe that with probability at least $1-2\cdot\exp\pth{-c_1\e^2k_2\rho(A^{*})/(1-p)^2}$:
\begin{equation}
\begin{split}
\sum_{i=1}^{k_1}\sigma_i^2+(1-\e')\sum_{i=k_1+1}^r \sigma_i^2 &\leq\|\tilde{A}\|_F^2\ \\ \leq & \sum_{i=1}^{k_1}\sigma_i^2+(1+\e')\sum_{i=k_1+1}^r\sigma_i^2  \label{eqn:frob-bd-simp1}
\end{split}
\end{equation}
Simplifying the upper bound of equation~\eqref{eqn:frob-bd-simp1} gives us,
\begin{eqnarray*}
 \|\tilde{A}\|_F^2 &\leq & \|A\|_F^2+\e'(\|A\|_F^2-\sum_{i=1}^{k_1}\sigma_i^2) 
                   \;\;=\;\; \|A\|_F^2(1+\e).
\end{eqnarray*}
Simplifying the lower bound of equation~\eqref{eqn:frob-bd-simp1} gives us $\|\tilde{A}\|_F^2 \geq \|A\|_F^2(1-\e)$.
Thus we get that 
\begin{eqnarray*}
    \Prob{\|A\|_F^2(1-\e) \leq \|\tilde{A}\|_F^2\leq \|A\|_F^2(1+\e)}  & \geq \\
    1-2\exp\pth{-\frac{c_1\e^2k_2\rho(A^{*})}{(1-p)^2}}.
\end{eqnarray*}
Now applying the definition of $\Lambda_{M_1}$, we finally get:
$$\Prob{\Lambda_{M1}\geq \e}\leq 2\cdot\exp\pth{-\frac{c_1\e^2k_2\rho(A^{*})}{(1-p)^2}}. \qedhere$$
\end{proof}
\section{Other proofs}
 \begin{thm}~\cite{HWineq} 
         \label{thm:hanson-wright-multi-dim}
         Let $G$ be a $D\times d$ random matrix with entries being independent gaussian random variables $G_{ij}$ with mean zero and variance $1/D$.
         Let $B$ be an $n\times n$ matrix with entries $b_{ij} \in \R$. Then for every $t\geq 0$, we have
         \begin{eqnarray*}
         \Prob{\left|\trace(GBG^\top)-\Ex{\trace(GBG^\top)} \right| \geq t} &\leq& 2\cdot\exp\pth{-c\cdot\min\pth{\frac{t^2}{\|B\|_F^2},\frac{t}{\|B\|}}}.
         \end{eqnarray*}
     \end{thm}
\begin{prf}[Proof of Lemma \ref{l:rand-map-avg-dist}]
     The main tool in our proof is the multi-dimensional variant of the \emph{Hanson-Wright inequality}~\cite{HWineq}, stated in Theorem \ref{thm:hanson-wright-multi-dim} which gives concentration bounds for certain quadratic forms of gaussian random variables.
    
         Let $Z := \|AG\|_F^2 = \trace(G^\top A^\top AG)$.
         We shall apply Theorem~\ref{thm:hanson-wright-multi-dim}, with $G$ as the $D\times d$ random matrix,
         and $B = A^\top A$ as a $D\times D$ matrix.
         We shall also require the following standard observations, which can be easily derived: 
         $\|B\| = \|A\|^2$, and $\|B\|_F^2 \leq \|A\|^2\cdot \|A\|_F^2$.
         This gives
         \begin{eqnarray*}
             \Prob{\left|Z-\Ex{Z} \right| \geq t} &\leq& 2\cdot\exp\pth{-c\cdot\min\pth{\frac{t^2}{D\|B\|_F^2},\frac{t}{\|B\|}}} \\
                                                  &\leq& 2\cdot\exp\pth{-c\cdot\min\pth{\frac{t^2}{D\|B\|_F^2},\frac{t}{\|A\|^2}}}. 
         \end{eqnarray*}
         By linearity of Expectation over Gaussian random variables, \\
         $\Ex{Z} = \Ex{ \trace(G^\top A^\top AG)} = \Ex{\trace(AGG^\top A^\top )} = d \cdot\trace(AA^\top )$ and thus,\\
         $\Ex{Z}=    d \sum_{r=1}^n a_r a_r^\top = \sum_{r=1}^n d\|a_r\|^2 = d\|A\|_F^2$\\
         Now taking $t= \e \Ex{Z} = \e d\|A\|_F^2$, we get
         \begin{eqnarray*}
             \Prob{\left|Z-d\|A\|_F^2\right| \geq \e d\|A\|_F^2} &\leq& 2\cdot\exp\pth{-c\cdot\min\pth{\frac{\e^2d^2\|A\|_F^4}{D\|A\|^2\|A\|_F^2},\frac{\e d\|A\|_F^2}{\|A\|^2}}} \\ 
             &=& 2\cdot\exp\pth{-c\cdot\min\pth{\frac{\e^2d^2\|A\|_F^2}{D\|A\|^2},\frac{\e d\|A\|_F^2}{\|A\|^2}}} \\ 
             &\leq& 2\cdot\exp\pth{-c\pth{\frac{\e^2 d\|A\|_F^2}{\|A\|^2}}}, 
         \end{eqnarray*}
         where the last line follows by observing that $d/D \leq 1$ for $d \leq D$. 
         $\qed$
     \end{prf}

\begin{prf}[Proof of Inequality \ref{stress-ineq} (Theorem \ref{stressBound})]
     We have:
$$a^2(c^2+b^2-2bc)=a^2(b-c)^2\geq 0 \implies a^2c^2+a^2b^2\geq 2a^2bc$$
$$a^4+a^2c^2+a^2b^2+b^2c^2\geq a^4+b^2c^2+2a^2bc = (a^2+bc)^2$$
Since $a,b,c$ are non negative, we can take a square root.
$$\sqrt{a^2+b^2}\sqrt{a^2+c^2}\geq a^2+bc$$
$$2a^2-2\sqrt{a^2+b^2}\sqrt{a^2+c^2}\leq -2bc$$
$$a^2+b^2+a^2+c^2-2\sqrt{a^2+b^2}\sqrt{a^2+c^2}\leq b^2+c^2-2bc$$
$$\left(\sqrt{a^2+b^2}-\sqrt{a^2+c^2}\right)^2\leq (b-c)^2  $$
    
\end{prf}
\begin{prf}[Proof of Lemma \ref{data-difference}]
     As the data is centered, the sum $\sum_{j=1}^n \mathbf{x}_j=0$. We have:
     $\sum_{j<i}^n ||\mathbf{d}_{ij}||^2 =
     n \sum_{i=1}^n ||\mathbf{x}_i||^2-\sum_{i=1}^{n}\mathbf{x}_i$
\end{prf}

\section{Complexity Analysis of the  \textit{DiffRed} Algorithm \ref{DiffRed-algo}}\label{sec:complexity}
\kunal{Based on the algorithm description given previously, the running time complexity of \textit{DiffRed} can be obtained as follows. We first obtain 
   a $k_1$-rank approximation of the $n\times D$ data matrix using the singular value decomposition. This takes $O(nD^2)$ time. Next, we generate and apply a random $k_2\times D$ Gaussian matrix, which can be done in time \prarabdh{$O(nk_2D)$. For $\eta$ Monte Carlo iterations, this becomes $O(\eta nk_2D)$.  Thus, the total time complexity comes to $O(nD^2 +\eta nk_2D)$. For the case when $D \gg n$, we work with $A^{\top}$, and thus get a complexity of $O(Dn^2+\eta nk_2D)$. So the overall complexity can be summarized as $O(Dn\cdot\min\{D,n\}+\eta nk_2D)$.} }
\section{Detailed Experiment Results [From Section \ref{sec:exp-res}]}
\subsection{\emph{Stress} and \emph{M1}: Datasetwise results}
In this section, we present a dataset wise summary of the results of our experiments on \emph{M1} and \emph{Stress} and in the later sections, we present the full grid search results. Figure \ref{fig:spectral-plots} shows the plots of the singular values of all the datasets. 
\input{supp_fig_files/spectral-plots}


Figure \ref{fig:sr-plots} shows how the stable rank of the residual matrix for all datasets increases as more directions of variance are removed (i.e., $k_1$) so long as the number of components removed remains well within the range of a practically required dimensionality ($<100$ for high dimensionality datasets). From our datasets, we note that for Bank and hatespeech, the starting dimensionality itself is low (17 and 100 respectively) and therefore the peak of the curve occurs earlier. 
\input{supp_fig_files/sr-plots}
\subsubsection{Bank}
Bank \cite{misc_bank_marketing_222} is a binary classification dataset of a Portugese bank's marketing campaign. The goal of the classification is to predict if a client will subscribe to a term deposit or not. The dataset has a low dimensionality of 17 which puts it out of the curse of dimensionality regime. We account for this low dimensionality in our experiments by exploring only low target dimensions in the range 1 to 8. It also has a relatively low stable rank of 1.48. Figure \ref{fig:bank-spectral} shows the singular value plot for Bank and Figure \ref{fig:bank-sr} shows the stable rank plot. Table \ref{tab:bank-dr-stress} and Table \ref{tab:bank-dr-m1} show the results of our grid search experiments on \emph{Stress} and \emph{M1} metrics to find the optimal values of $k_1$ and $k_2$. The rows having the minimum metric value are marked in bold. 
\vspace*{-\baselineskip}
\subsubsection{Hatespeech}
Hatespeech \cite{hateoffensive} is a dataset of tweets labelled according to different types of offensive content. For our experiments we use the \texttt{.npy} files provided by \cite{Espadoto2019TowardAQ} on their paper website \footnote{\href{https://mespadoto.github.io/proj-quant-eval/}{https://mespadoto.github.io/proj-quant-eval/}}. The dataset has a dimensionality of 100 and a stable rank of 11. Figure \ref{fig:hatespeech-spectral} shows the plot of singular values and Figure \ref{fig:hatespeech-sr} shows the stable rank plot. Tables \ref{tab:hatespeech-dr-stress} and \ref{tab:hatespeech-dr-m1} show the results of our experiments on \emph{M1} and \emph{Stress}. 
\vspace*{-\baselineskip}

\subsubsection{FMnist}
Fashion-MNIST \cite{xiao2017fashionmnist} (abbreviated as FMnist in our paper) is an image dataset consisting of 60,000 images of fashion images belonging to 10 different classes. Each image is a 28 by 28 grayscale image which can be represented as a 784 dimensional vector. The dataset has a stable rank of 2.68. Figure \ref{fig:hatespeech-spectral} shows the singular value plot and Figure \ref{fig:hatespeech-sr} shows the stable rank plot for FMnist. Tables \ref{tab:fmnist-dr-stress} and \ref{tab:fmnist-dr-m1} show our experiment results. 
\vspace*{-\baselineskip}
\subsubsection{Cifar10}
Cifar10 \cite{Krizhevsky2009LearningML} is a dataset of 60,000 color images belonging to 10 classes. Each image is a 32 by 32 image with 3 channels, therefore, each image can be represented as a 3072 dimensionality vector. The stable rank of the dataset is 6.13. Figures \ref{fig:Cifar10-spectral} and \ref{fig:Cifar10-sr} are the relevant spectral and stable rank plots. Tables \ref{tab:cifar-dr-stress} and \ref{tab:cifar-dr-m1} show the results of our experiments on \emph{Stress} and \emph{M1}. 
\vspace*{-\baselineskip}
\subsubsection{geneRNASeq}
geneRNASeq \cite{misc_gene_expression_cancer_rna-seq_401} is a dataset of gene expressions with the aim of classifying 5 types of tumor. It has a dimensionality of 20531 and a stable rank of 1.12 which is also the lowest stable rank among all datasets. Figures \ref{fig:geneRNASeq-spectral} and \ref{fig:geneRNASeq-sr} show the spectral and stable rank plots respectively. Tables \ref{tab:gene-dr-stress} and \ref{tab:gene-dr-m1} show the results of our grid search experiments on $k_1$ and $k_2$.
\vspace*{-\baselineskip}
\subsubsection{Reuters30k}
Reuters30k is a TF-IDF vector respresentation of the Reuters newswires dataset \cite{misc_reuters-21578_text_categorization_collection_137} which is a collection of news articles belonging to different topics. We use this\footnote{\href{https://huggingface.co/datasets/reuters21578}{https://huggingface.co/datasets/reuters21578}} huggingface version of the dataset. To generate TF-IDF representation, we use scikit-learn's \texttt{TfidVectorizer}. The dimensionality of the dataset becomes 30,916 after this preprocessing. It also has the highest stable rank (14.50) among all datasets. Figures \ref{fig:Reuters30k-spectral} and \ref{fig:Reuters30k-sr} show the relevant spectral and stable rank plots. Tables \ref{tab:reuters-dr-stress} and \ref{tab:reuters-dr-m1} show the results of our experiments on \emph{Stress} and \emph{M1}. Another interesting observation is that for target dimension 40, we achieve 81.87\% reduction in \emph{Stress} as compared to PCA (marked in red). 
\subsection{Very High Dimensionality Datasets}
We chose two very high dimensionality datasets- APTOS 2019 (509k) and DIV2k (6.6M)- one with a low stable rank and one with a high stable rank. We only evaluated these datasets on PCA and RMap other than \textit{DiffRed} because their high dimensionality made other algorithms very slow. 
\input{supp_fig_files/hdd-spectral-plots}
\input{supp_fig_files/hdd-sr-plots}
\subsubsection{APTOS 2019}
APTOS 2019 \cite{aptos2019-blindness-detection} is a Kaggle dataset of 13,000 retina images taken using fundus photography. It is a multiclass-classification dataset where each image is labelled as belonging to one of the five levels of severity of diabetic retinopathy. For our purpose, we resized each image to size of 474 by 358 yielding vectors of dimensionality 509,076 (as each image has 3 channels). This is one of our datasets in the 'very high dimensionality' category. It has a low stable rank of 1.32. Figures \ref{fig:aptos2019-spectral} and \ref{fig:aptos2019-sr} show the spectral and stable rank plots for APTOS 2019 and Tables \ref{tab:aptos-dr-stress} and \ref{tab:aptos-dr-m1} show the results of the grid search experiments on $k_1$ and $k_2$ for \emph{Stress} and \emph{M1} metrics. 
\subsubsection{DIV2k}
DIV2k \cite{8014884,Agustsson_2017_CVPR_Workshops} is a dataset of 800 2K high resolution image dataset from the NTIRE 2017 challenge. For our purposes, we rescale every image to 1080 by 2048 which means that each image can be represented as a 6,635,520 dimensional vector (3 channels of color). The dataset has a high stable rank of 8.39. Figures \ref{fig:div2k-spectral} and \ref{fig:div2k-sr} are the respective spectral and stable rank plots. Tables \ref{tab:div2k-dr-stress} and \ref{tab:div2k-dr-m1} show the results of our grid search experiments to find optimal $k_1$ and $k_2$ for \emph{Stress} and \emph{M1} metrics. 
\subsection{Low sensitivity of $\Lambda_{M_1}$ to $k_1$ and $k_2$}
The following table shows the Average Variance of $\Lambda_{M_1}$ over different $k_1$ and $k_2$ values for different dimensions (described in Section \ref{sec:M1}, \textbf{Observation 2} ). 
\input{tables/m1-diffred-avg_var}
\subsection{Hyperparameter tuning}
For our experiments on hyperparameter tuning for other dimensionality reduction techniques, we have presented the most optimal values of \emph{Stress} and \emph{M1} in our tables. For full results, please refer to the files in the following directories in our repository\footnote{\href{https://github.com/S3-Lab-IIT/DiffRed}{\texttt{\textcolor{blue}{https://github.com/S3-Lab-IIT/DiffRed}}}}:
\begin{itemize}
    \item Full results: \texttt{Experiments/dimensionality\_reduction\_metrics/results/other\_dr\_techniques/}
    \item Code: \texttt{Experiments/dimensionality\_reduction\_metrics/other\_dr\_techniques/}
\end{itemize}
\subsection{Effect of Monte Carlo iterations [From Section \ref{sec:Discusssion}]}
Figure \ref{fig:mc-plots} shows that performing Monte Carlo iterations helps in finding good random directions. With more Monte Carlo iterations, we find Random Maps that further reduce the \emph{M1} metric. We note that such Random Maps (which minimize \emph{M1}), also further reduce \emph{Stress}. All the results presented in the paper have been computed at $\eta=100$.

\input{supp_fig_files/mc-plots}

\subsection{Application: \textit{DiffRed} as a precursor to visualization}

PCA has been widely used to reduce the dimensionality of high-dimensional datasets before applying T-SNE/UMap to mitigate the slow computation for high dimensions. Using \textit{DiffRed}, we can reduce the data to an intermediate dimension while preserving \emph{Stress} (global structure) and then apply T-SNE/UMap for visualization. The following Table \ref{tab:dr-vis-results} shows that using \textit{DiffRed} as a preprocessing step causes significant improvement in the \emph{Stress} of the final T-SNE/UMap visualization for the Reuters30k dataset.
\begin{table}[h]
\centering
\begin{tabular}{|l|c|}
\hline
\multicolumn{1}{|c|}{Method} & $\Lambda_S$   \\ \hline
PCA + T-SNE                  & 0.55          \\ \hline
\textit{DiffRed} + T-SNE              & \textbf{0.32} \\ \hline
PCA + UMap                   & 0.56          \\ \hline
\textit{DiffRed} + UMap               & \textbf{0.45} \\ \hline
\end{tabular}

\caption{\emph{Stress} of T-SNE and UMap after using PCA \& \textit{DiffRed} as pre-processing step for Reuters30k with intermediate dimension 10. Final \emph{Stress} for the T-SNE2 and UMap2 versions described in the main paper are presented here.}
\label{tab:dr-vis-results}
\end{table}
\squeezeup
\vspace*{-\baselineskip}
\section{Tables} \label{sec:tables}
In this section, we provide data from various hyper-parameter tuning experiments. For \textit{DiffRed}, we varied the target dimension and the values of $k_1$ and $k_2$. It is clear from these experiments, that by increasing the target dimension, we can reduce the \emph{Stress} metric. The \emph{M1} metric is not sensitive to the choice of $k_1$ and $k_2$. However, the values of $k_1$ and $k_2$ have to be chosen carefully to minimize \emph{Stress}. The optimal choice can be made by using the theoretical bound as discussed in the main text of the paper in Section \ref{sec:stress}.

\subsection{\emph{Stress}}
\input{supp_tables/bank-dr-stress}
\input{supp_tables/hatespeech-dr-stress}
\input{supp_tables/fmnist-dr-stress}
\input{supp_tables/cifar-dr-stress}
\input{supp_tables/gene-dr-stress}
\input{supp_tables/reuters-dr-stress}
\input{supp_tables/aptos-dr-stress}
\input{supp_tables/div2k-dr-stress}

\subsection{\emph{M1}}
\squeezeup
\vspace*{-\baselineskip}
\input{supp_tables/bank-dr-m1}
\input{supp_tables/hatespeech-dr-m1}
\input{supp_tables/fmnist-dr-m1}
\squeezeup
\vspace*{-\baselineskip}
\input{supp_tables/cifar-dr-m1}
\squeezeup
\vspace*{-\baselineskip}
\input{supp_tables/gene-dr-m1}
\squeezeup
\vspace*{-\baselineskip}
\input{supp_tables/reuters-dr-m1}
\squeezeup
\vspace*{-\baselineskip}
\input{supp_tables/aptos-dr-m1}
\input{supp_tables/div2k-dr-m1}
\vfill

%% file: supp_fig_files/spectral-plots.tex
\begin{figure}[H]
    \centering %
    \captionsetup{font=small}
    \begin{subfigure}[t]{0.35\textwidth}
    \centering
    \includegraphics[width=\textwidth]{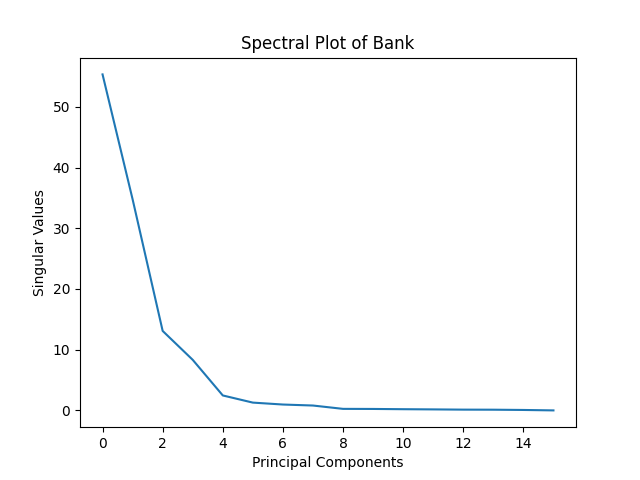}
    \caption{Bank}
    \label{fig:bank-spectral}
\end{subfigure}%
~
 \begin{subfigure}[t]{0.35\textwidth}
    \centering
    \includegraphics[width=\textwidth]{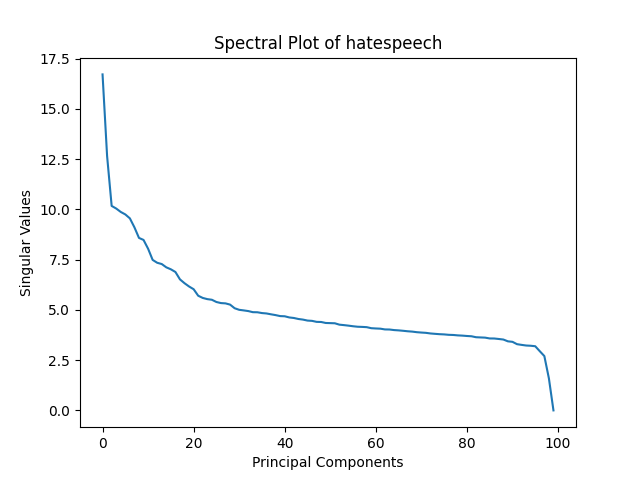}
    \caption{Hatespeech}
    \label{fig:hatespeech-spectral}
\end{subfigure}
 \begin{subfigure}[t]{0.35\textwidth}
    \centering
    \includegraphics[width=\textwidth]{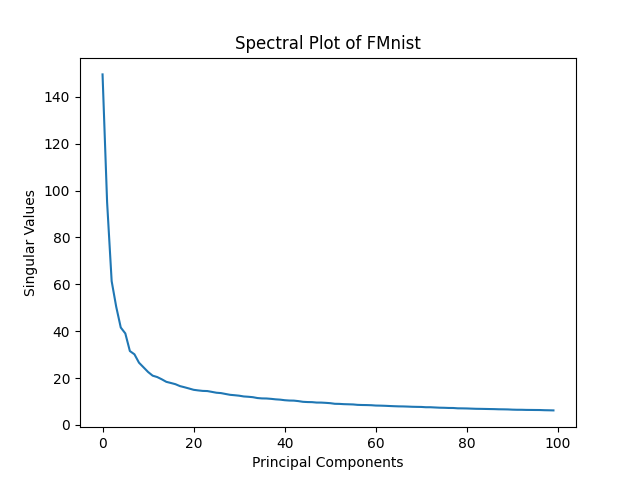}
    \caption{FMnist}
    \label{fig:FMnist-spectral}
\end{subfigure}%
\begin{subfigure}[t]{0.35\textwidth}
    \centering
    \includegraphics[width=\textwidth]{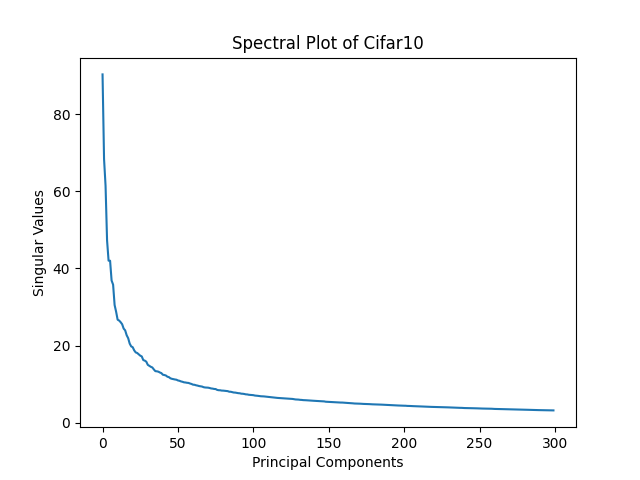}
    \caption{Cifar10}
    \label{fig:Cifar10-spectral}
\end{subfigure}
\begin{subfigure}[t]{0.35\textwidth}
    \centering
    \includegraphics[width=\textwidth]{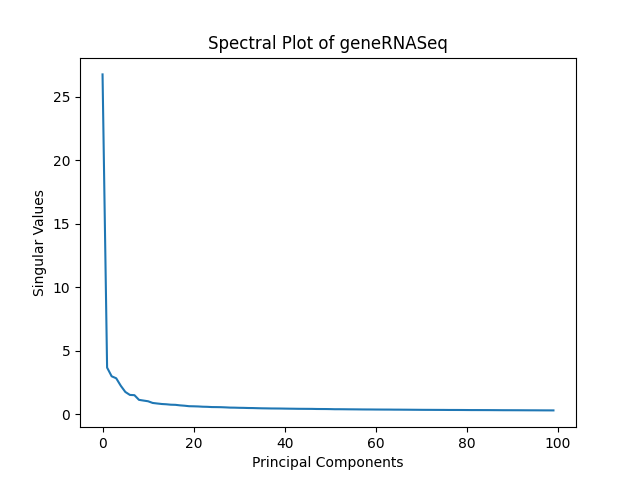}
    \caption{geneRNASeq}
    \label{fig:geneRNASeq-spectral}
\end{subfigure}%
\begin{subfigure}[t]{0.35\textwidth}
    \centering
    \includegraphics[width=\textwidth]{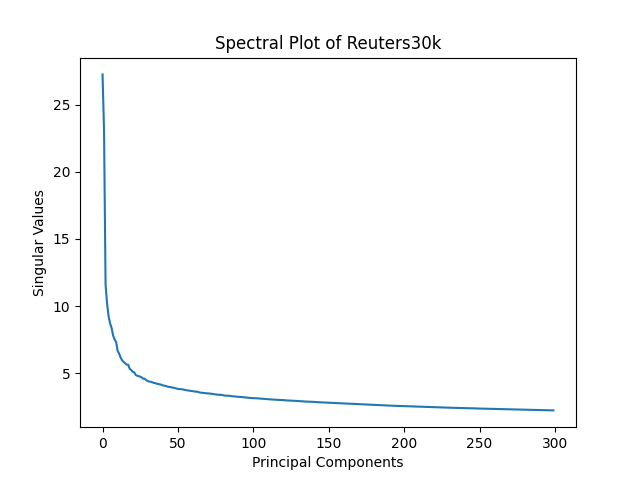}
    \caption{Reuters30k}
    \label{fig:Reuters30k-spectral}
\end{subfigure}
\caption{Plots showing the spectral plots of all the datasets}
\label{fig:spectral-plots}
\end{figure}

%% file: supp_fig_files/sr-plots.tex
\begin{figure}[H]
    \centering %
    \captionsetup{font=small}
    \begin{subfigure}[t]{0.45\textwidth}
    \centering
    \includegraphics[width=\textwidth]{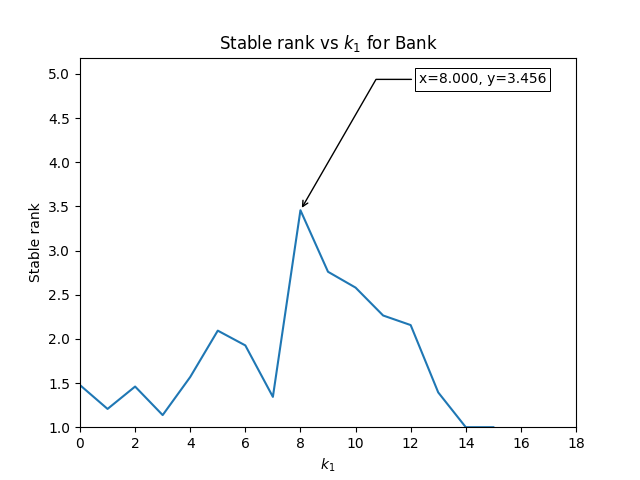}
    \caption{Bank}
    \label{fig:bank-sr}
\end{subfigure}%
~
 \begin{subfigure}[t]{0.45\textwidth}
    \centering
    \includegraphics[width=\textwidth]{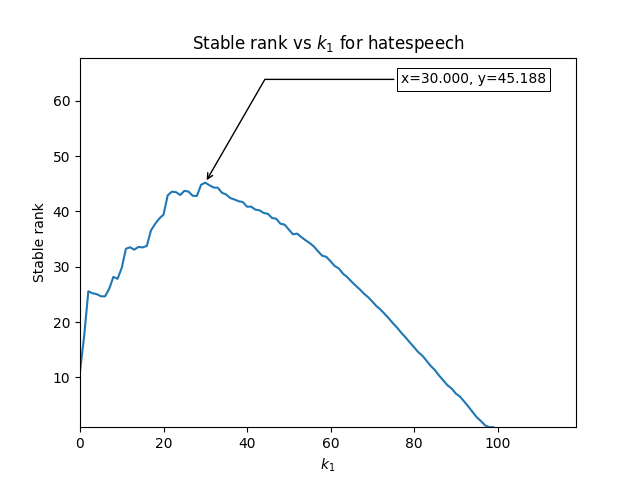}
    \caption{Hatespeech}
    \label{fig:hatespeech-sr}
\end{subfigure}
 \begin{subfigure}[t]{0.45\textwidth}
    \centering
    \includegraphics[width=\textwidth]{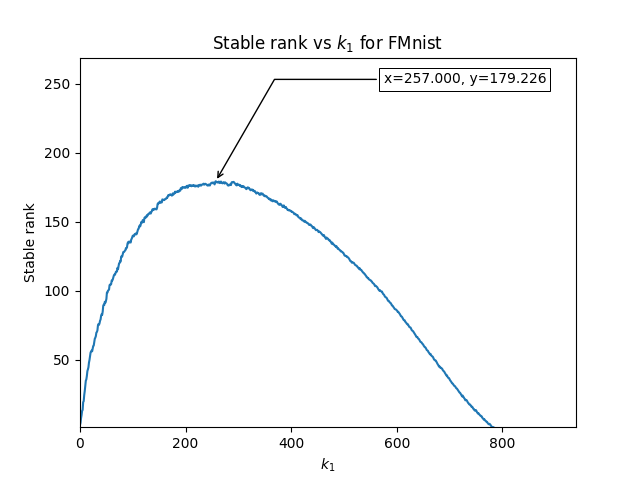}
    \caption{FMnist}
    \label{fig:FMnist-sr}
\end{subfigure}%
\begin{subfigure}[t]{0.45\textwidth}
    \centering
    \includegraphics[width=\textwidth]{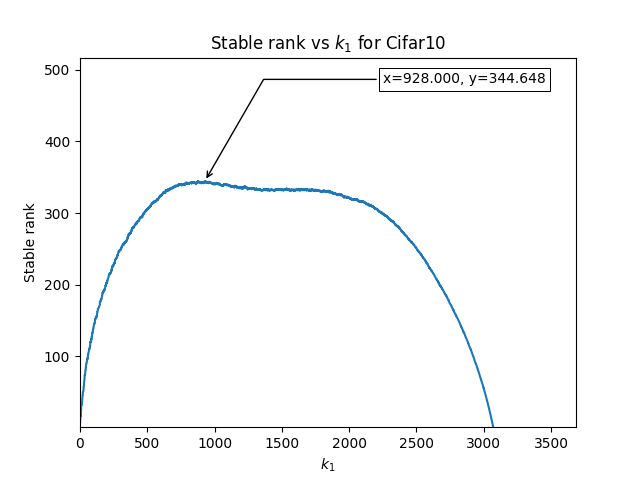}
    \caption{Cifar10}
    \label{fig:Cifar10-sr}
\end{subfigure}
\begin{subfigure}[t]{0.45\textwidth}
    \centering
    \includegraphics[width=\textwidth]{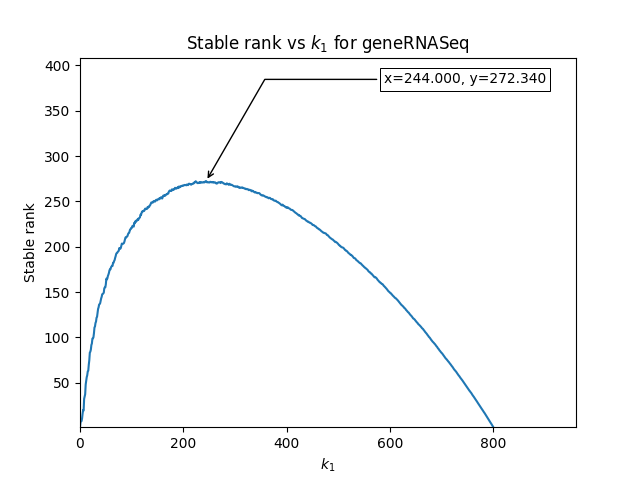}
    \caption{geneRNASeq}
    \label{fig:geneRNASeq-sr}
\end{subfigure}%
\begin{subfigure}[t]{0.45\textwidth}
    \centering
    \includegraphics[width=\textwidth]{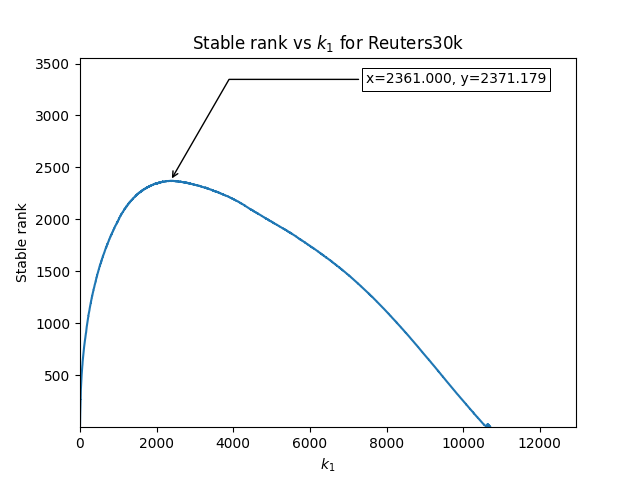}
    \caption{Reuters30k}
    \label{fig:Reuters30k-sr}
\end{subfigure}
\caption{Plots showing the stable rank vs. $k_1$ for all the datasets}
\label{fig:sr-plots}
\end{figure}

%% file: supp_fig_files/hdd-spectral-plots.tex
\begin{figure}
    \centering
    \begin{subfigure}[t]{0.35\textwidth}
    \centering
    \includegraphics[scale=0.35]{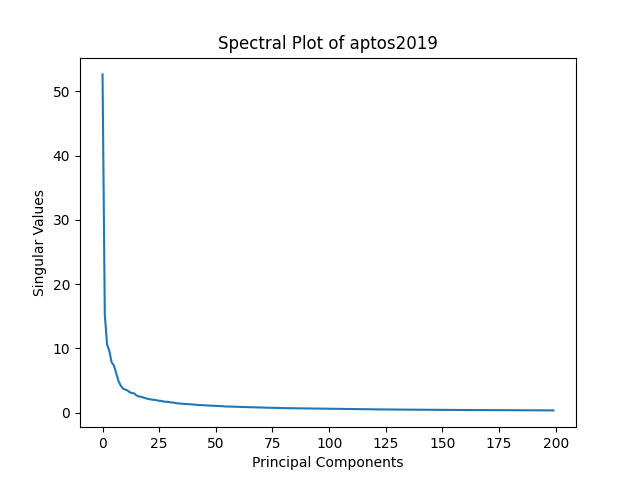}
    \caption{APTOS 2019}
    \label{fig:aptos2019-spectral}
\end{subfigure}%
~
\begin{subfigure}[t]{0.35\textwidth}
    \centering
    \includegraphics[scale=0.35]{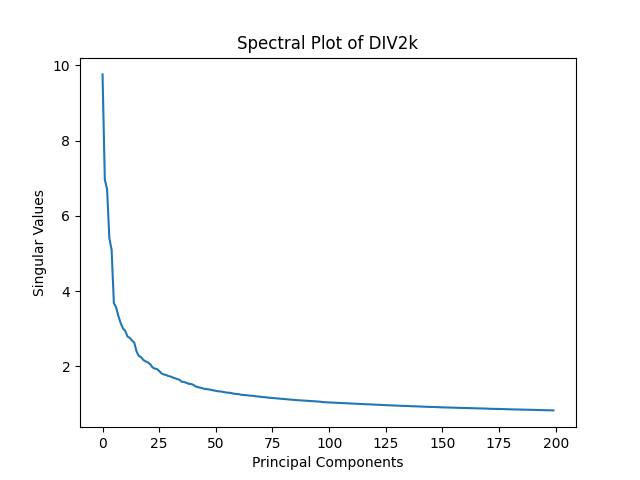}
    \caption{DIV2k}
    \label{fig:div2k-spectral}
\end{subfigure}
\caption{Spectral plots of very high dimensionality datasets.}
\label{fig:hdd-spectral-plots}
\end{figure}

%% file: supp_fig_files/hdd-sr-plots.tex
\begin{figure}
    \centering
    \begin{subfigure}[t]{0.35\textwidth}
    \centering
    \includegraphics[scale=0.35]{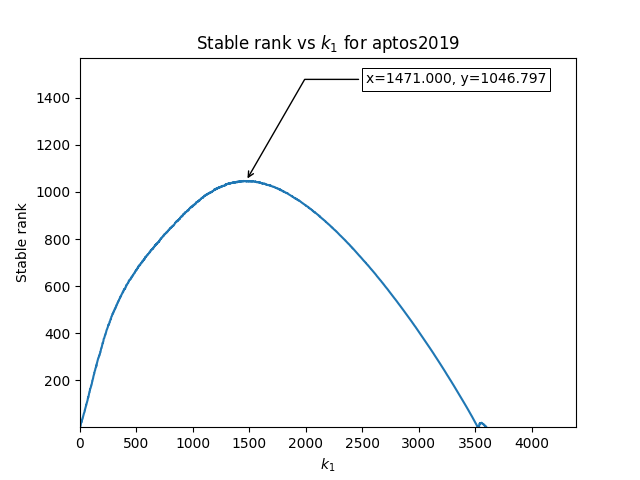}
    \caption{APTOS 2019}
    \label{fig:aptos2019-sr}
\end{subfigure}%
~
\begin{subfigure}[t]{0.35\textwidth}
    \centering
    \includegraphics[scale=0.35]{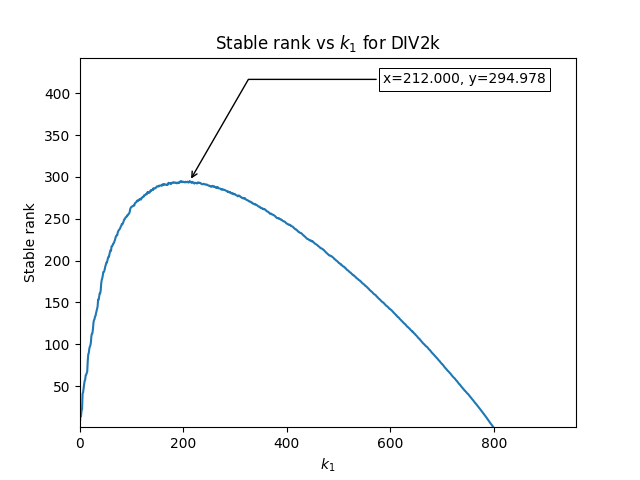}
    \caption{DIV2k}
    \label{fig:div2k-sr}
\end{subfigure}
\caption{Stable Rank plots of very high dimensionality datasets.}
\label{fig:hdd-sr-plots}
\end{figure}

%% file: tables/m1-diffred-avg_var.tex
\begin{table}[H]
\centering
\captionsetup{font=footnotesize}
\begin{tabular}{|l|c|}
\hline
\textbf{Dataset} & $\beta$ \\ \hline
Bank             & 3.85e-06                  \\
Hatespeech       & 3.45e-07                  \\
FMnist           & 9.85e-07                  \\
Cifar10          & 3.25e-07                  \\
geneRNASeq       & 3.59e-06                  \\
Reuters30k       & 2.17e-08                  \\
APTOS 2019       & 3.07e-06                  \\
DIV2k            & 9.80e-08                  \\ \hline
\end{tabular}
\caption{Variance $\beta$ (defined in Sec. \ref{sec:M1}, \textbf{Observation 2}) observed in $\Lambda_{M_1}$ for different combinations of $k_1$ and $k_2$ averaged over all target dimensions.}
\label{tab:m1-diffred-avg_var}
\end{table}

%% file: supp_fig_files/mc-plots.tex
\begin{figure}[H]
    \centering
    \begin{subfigure}[t]{0.35\textwidth}
    \centering
    \includegraphics[width=\textwidth]{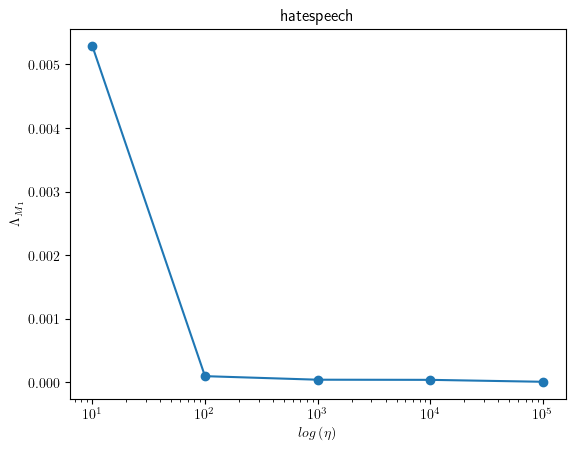}
    \caption{M1}
    \label{fig:mc-m1}
\end{subfigure}%
~
\begin{subfigure}[t]{0.35\textwidth}
    \centering
    \includegraphics[width=\textwidth]{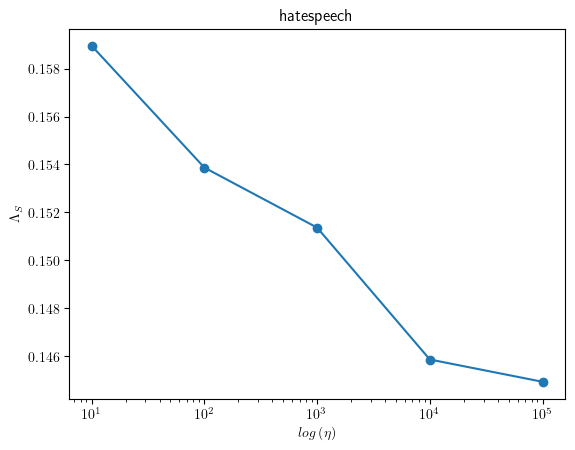}
    \caption{Stress}
    \label{fig:mc-stress}
\end{subfigure}
\caption{Log scale plot of Stress and M1 metrics against the number of Monte Carlo iterations $\eta$ showing how diminishing improvements over the metrics are obtained with increasing $\eta$}
\label{fig:mc-plots}
\end{figure}

%% file: supp_tables/bank-dr-stress.tex
\squeezeup
\vspace*{-\baselineskip}
\begin{adjustwidth}{-2.5 cm}{-2.5 cm}\centering
\resizebox{\columnwidth}{!}{
\begin{threeparttable}[t]
\caption{Bank:$\Lambda_S$}\label{tab:bank-dr-stress}
\scriptsize
\begin{tabular}{ccccccccccc}\toprule\toprule
\multirow{2}{*}{\textbf{Target Dimension}} &\multirow{2}{*}{\textbf{k1}} &\multirow{2}{*}{\textbf{k2}} &\multirow{2}{*}{\textbf{Stress}} &\multirow{2}{*}{\textbf{PCA Stress}} &\multicolumn{2}{c}{\textbf{RMap Stress ($\alpha$=20)}} &\multirow{2}{*}{\textbf{S-PCA Stress}} &\multirow{2}{*}{\textbf{K-PCA Stress}} &\multirow{2}{*}{\textbf{UMAP Stress}} \\\cmidrule{6-7}
& & & & &$\mathbf{\mu}$ &$\mathbf{\sigma}$ & & & \\\midrule
1 &0 &1 &0.278474 &0.175070 &0.40 &0.128 &0.189710 &0.507601 &12.799170 \\
\specialrule{0.3pt}{0.1pt}{0.1pt}
2 &0 &2 &0.413185 &\multirow{2}{*}{0.085241} &\multirow{2}{*}{0.28} &\multirow{2}{*}{0.095} &\multirow{2}{*}{0.099255} &\multirow{2}{*}{0.481864} &\multirow{2}{*}{8.930940} \\
\textbf{2} &\textbf{1} &\textbf{1} &\textbf{0.310159} & & & & & & \\
\specialrule{0.3pt}{0.1pt}{0.1pt}
3 &0 &3 &0.250078 &\multirow{3}{*}{0.038763} &\multirow{3}{*}{0.28} &\multirow{3}{*}{0.060} &\multirow{3}{*}{0.057670} &\multirow{3}{*}{0.470851} &\multirow{3}{*}{7.912646} \\
3 &1 &2 &0.169713 & & & & & & \\
\textbf{3} &\textbf{2} &\textbf{1} &\textbf{0.056258} & & & & & & \\
\specialrule{0.3pt}{0.1pt}{0.1pt}
5 &0 &5 &0.117427 &\multirow{5}{*}{0.003634} &\multirow{5}{*}{0.22} &\multirow{5}{*}{0.098} &\multirow{5}{*}{0.036128} &\multirow{5}{*}{0.465844} &\multirow{5}{*}{7.402985} \\
5 &1 &4 &0.098115 & & & & & & \\
5 &2 &3 &0.050027 & & & & & & \\
5 &3 &2 &0.010721 & & & & & & \\
\textbf{5} &\textbf{4} &\textbf{1} &\textbf{0.004978} & & & & & & \\
\specialrule{0.3pt}{0.1pt}{0.1pt}
6 &0 &6 &0.082337 &\multirow{6}{*}{0.003634} &\multirow{6}{*}{0.17} &\multirow{6}{*}{0.059} &\multirow{6}{*}{0.037417} &\multirow{6}{*}{0.465480} &\multirow{6}{*}{7.069356} \\
6 &1 &5 &0.073995 & & & & & & \\
6 &2 &4 &0.02745 & & & & & & \\
6 &3 &3 &0.012201 & & & & & & \\
6 &4 &2 &0.004184 & & & & & & \\
\textbf{6} &\textbf{5} &\textbf{1} &\textbf{0.00235} & & & & & & \\
\specialrule{0.3pt}{0.1pt}{0.1pt}
7 &0 &7 &0.1341 &\multirow{7}{*}{0.001219} &\multirow{7}{*}{0.17} &\multirow{7}{*}{0.091} &\multirow{7}{*}{0.036316} &\multirow{7}{*}{0.465270} &\multirow{7}{*}{7.107705} \\
7 &1 &6 &0.053662 & & & & & & \\
7 &2 &5 &0.011971 & & & & & & \\
7 &3 &4 &0.00538 & & & & & & \\
7 &4 &3 &0.002543 & & & & & & \\
7 &5 &2 &0.00159 & & & & & & \\
\textbf{7} &\textbf{6} &\textbf{1} &\textbf{0.00109} & & & & & & \\
\specialrule{0.3pt}{0.1pt}{0.1pt}
8 &0 &8 &0.109656 &\multirow{7}{*}{0.000674} &\multirow{7}{*}{0.18} &\multirow{7}{*}{0.070} &\multirow{7}{*}{0.037008} &\multirow{7}{*}{0.465107} &\multirow{7}{*}{7.131875} \\
8 &2 &6 &0.032214 & & & & & & \\
8 &3 &5 &0.007498 & & & & & & \\
8 &4 &4 &0.002741 & & & & & & \\
8 &5 &3 &0.002032 & & & & & & \\
8 &6 &2 &0.00073 & & & & & & \\
\textbf{8} &\textbf{7} &\textbf{1} &\textbf{0.000424} & & & & & & \\
\bottomrule
\end{tabular}
\end{threeparttable}
}
\end{adjustwidth}

%% file: supp_tables/hatespeech-dr-stress.tex
\begin{adjustwidth}{-2.5 cm}{-2.5 cm}\centering
\resizebox{0.95\columnwidth}{!}{
\begin{threeparttable}
\caption{Hatespeech: $\Lambda_S$}\label{tab:hatespeech-dr-stress}
\scriptsize
\begin{tabular}{ccccccccccc}\toprule\toprule
\multirow{2}{*}{\textbf{Target Dimension}} &\multirow{2}{*}{\textbf{k1}} &\multirow{2}{*}{\textbf{k2}} &\multirow{2}{*}{\textbf{Stress}} &\multirow{2}{*}{\textbf{PCA Stress}} &\multicolumn{2}{c}{\textbf{RMap Stress ($\alpha$=20)}} &\multirow{2}{*}{\textbf{S-PCA Stress}} &\multirow{2}{*}{\textbf{K-PCA Stress}} &\multirow{2}{*}{\textbf{UMAP Stress}} \\\cmidrule{6-7}
& & & & &$\mathbf{\mu}$ &$\mathbf{\sigma}$ & & & \\\midrule
10 &0 &10 &0.154463 &\multirow{8}{*}{0.36} &\multirow{8}{*}{0.16} &\multirow{8}{*}{0.01} &\multirow{8}{*}{0.36} &\multirow{8}{*}{0.65} &\multirow{8}{*}{5.29} \\
\textbf{10} &\textbf{1} &\textbf{9} &\textbf{0.152249} & & & & & & \\
10 &2 &8 &0.161939 & & & & & & \\
10 &3 &7 &0.159242 & & & & & & \\
10 &4 &6 &0.167399 & & & & & & \\
10 &5 &5 &0.167598 & & & & & & \\
10 &6 &4 &0.182932 & & & & & & \\
10 &7 &3 &0.207632 & & & & & & \\
\specialrule{0.3pt}{0.1pt}{0.1pt}
20 &0 &20 &0.108961 &\multirow{10}{*}{0.26} &\multirow{10}{*}{0.11} &\multirow{10}{*}{0.00} &\multirow{10}{*}{0.27} &\multirow{10}{*}{0.64} &\multirow{10}{*}{5.21} \\
20 &2 &18 &0.107507 & & & & & & \\
20 &3 &17 &0.098006 & & & & & & \\
20 &4 &16 &0.098426 & & & & & & \\
\textbf{20} &\textbf{5} &\textbf{15} &\textbf{0.097116} & & & & & & \\
20 &8 &12 &0.099934 & & & & & & \\
20 &10 &10 &0.107269 & & & & & & \\
20 &12 &8 &0.115551 & & & & & & \\
20 &15 &5 &0.127951 & & & & & & \\
20 &18 &2 &0.181814 & & & & & & \\
\specialrule{0.3pt}{0.1pt}{0.1pt}
30 &0 &30 &0.089438 &\multirow{13}{*}{0.20} &\multirow{13}{*}{0.09} &\multirow{13}{*}{0.00} &\multirow{13}{*}{0.22} &\multirow{13}{*}{0.63} &\multirow{13}{*}{5.17} \\
30 &2 &28 &0.085617 & & & & & & \\
30 &3 &27 &0.083632 & & & & & & \\
30 &5 &25 &0.079905 & & & & & & \\
30 &6 &24 &0.082452 & & & & & & \\
30 &8 &22 &0.076772 & & & & & & \\
30 &10 &20 &0.07936 & & & & & & \\
\textbf{30} &\textbf{12} &\textbf{18} &\textbf{0.073679} & & & & & & \\
30 &15 &15 &0.076703 & & & & & & \\
30 &18 &12 &0.086769 & & & & & & \\
30 &20 &10 &0.090678 & & & & & & \\
30 &25 &5 &0.103606 & & & & & & \\
30 &27 &3 &0.127849 & & & & & & \\
\specialrule{0.3pt}{0.1pt}{0.1pt}
40 &0 &40 &0.078129 &\multirow{13}{*}{0.16} &\multirow{13}{*}{0.08} &\multirow{13}{*}{0.00} &\multirow{13}{*}{0.17} &\multirow{13}{*}{0.63} &\multirow{13}{*}{5.09} \\
40 &2 &38 &0.071711 & & & & & & \\
40 &4 &36 &0.07244 & & & & & & \\
40 &5 &35 &0.069092 & & & & & & \\
40 &8 &32 &0.064041 & & & & & & \\
40 &10 &30 &0.064503 & & & & & & \\
40 &11 &29 &0.066805 & & & & & & \\
40 &15 &25 &0.059473 & & & & & & \\
40 &16 &24 &0.058154 & & & & & & \\
\textbf{40} &\textbf{20} &\textbf{20} &\textbf{0.058028} & & & & & & \\
40 &25 &15 &0.062881 & & & & & & \\
40 &30 &10 &0.071768 & & & & & & \\
40 &35 &5 &0.086825 & & & & & & \\
\bottomrule
\end{tabular}
\end{threeparttable}
}
\end{adjustwidth}

%% file: supp_tables/fmnist-dr-stress.tex
\begin{adjustwidth}{-2.5 cm}{-2.5 cm}\centering
\resizebox{0.98\columnwidth}{!}{
\begin{threeparttable}[p]
\caption{FMnist: $\Lambda_S$}\label{tab:fmnist-dr-stress}
\scriptsize
\begin{tabular}{ccccccccccc}\toprule\toprule
\multirow{2}{*}{\textbf{Target Dimension}} &\multirow{2}{*}{\textbf{k1}} &\multirow{2}{*}{\textbf{k2}} &\multirow{2}{*}{\textbf{Stress}} &\multirow{2}{*}{\textbf{PCA Stress}} &\multicolumn{2}{c}{\textbf{RMap Stress ($\alpha$=20)}} &\multirow{2}{*}{\textbf{S-PCA Stress}} &\multirow{2}{*}{\textbf{K-PCA Stress}} &\multirow{2}{*}{\textbf{UMAP Stress}} \\\cmidrule{6-7}
& & & & &$\mathbf{\mu}$ &$\mathbf{\sigma}$ & & & \\\midrule
10 &0 &10 &0.149077 &\multirow{7}{*}{0.19} &\multirow{7}{*}{0.15} &\multirow{7}{*}{0.009} &\multirow{7}{*}{0.21} &\multirow{7}{*}{0.68} &\multirow{7}{*}{4.02} \\
10 &2 &8 &0.12508 & & & & & & \\
10 &3 &7 &0.117871 & & & & & & \\
10 &4 &6 &0.120748 & & & & & & \\
\textbf{10} &\textbf{5} &\textbf{5} &\textbf{0.117036} & & & & & & \\
10 &6 &4 &0.124043 & & & & & & \\
10 &7 &3 &0.134036 & & & & & & \\
\specialrule{0.3pt}{0.1pt}{0.1pt}
20 &0 &20 &0.111262 &\multirow{10}{*}{0.14} &\multirow{10}{*}{0.11} &\multirow{10}{*}{0.009} &\multirow{10}{*}{0.16} &\multirow{10}{*}{0.68} &\multirow{10}{*}{4.34} \\
20 &2 &18 &0.085441 & & & & & & \\
20 &4 &16 &0.073317 & & & & & & \\
20 &5 &15 &0.068579 & & & & & & \\
20 &6 &14 &0.069169 & & & & & & \\
\textbf{20} &\textbf{8} &\textbf{12} &\textbf{0.066389} & & & & & & \\
20 &10 &10 &0.067899 & & & & & & \\
20 &12 &8 &0.070051 & & & & & & \\
20 &15 &5 &0.080917 & & & & & & \\
20 &18 &2 &0.112871 & & & & & & \\
\specialrule{0.3pt}{0.1pt}{0.1pt}
30 &0 &30 &0.095465 &\multirow{11}{*}{0.12} &\multirow{11}{*}{0.09} &\multirow{11}{*}{0.008} &\multirow{11}{*}{0.14} &\multirow{11}{*}{0.68} &\multirow{11}{*}{4.98} \\
30 &3 &27 &0.06347 & & & & & & \\
30 &5 &25 &0.053878 & & & & & & \\
30 &8 &22 &0.048813 & & & & & & \\
30 &10 &20 &0.047958 & & & & & & \\
30 &12 &18 &0.048057 & & & & & & \\
\textbf{30} &\textbf{15} &\textbf{15} &\textbf{0.047662} & & & & & & \\
30 &18 &12 &0.050517 & & & & & & \\
30 &20 &10 &0.052338 & & & & & & \\
30 &25 &5 &0.067108 & & & & & & \\
30 &27 &3 &0.083181 & & & & & & \\
\specialrule{0.3pt}{0.1pt}{0.1pt}
40 &0 &40 &0.085391 &\multirow{11}{*}{0.10} &\multirow{11}{*}{0.08} &\multirow{11}{*}{0.006} &\multirow{11}{*}{0.13} &\multirow{11}{*}{0.68} &\multirow{11}{*}{4.18} \\
40 &4 &36 &0.048749 & & & & & & \\
40 &5 &35 &0.045706 & & & & & & \\
40 &8 &32 &0.040424 & & & & & & \\
40 &10 &30 &0.039411 & & & & & & \\
\textbf{40} &\textbf{15} &\textbf{25} &\textbf{0.037091} & & & & & & \\
40 &16 &24 &0.037105 & & & & & & \\
40 &20 &20 &0.037701 & & & & & & \\
40 &25 &15 &0.039636 & & & & & & \\
40 &30 &10 &0.045351 & & & & & & \\
40 &35 &5 &0.058315 & & & & & & \\
\bottomrule
\end{tabular}
\end{threeparttable}
}
\end{adjustwidth}

%% file: supp_tables/cifar-dr-stress.tex
\begin{adjustwidth}{-2.5 cm}{-2.5 cm}
\centering
\resizebox{0.98\columnwidth}{!}{
\begin{threeparttable}[p]
\caption{Cifar10: $\Lambda_S$}\label{tab:cifar-dr-stress}
\scriptsize
\begin{tabular}{ccccccccccc}\toprule\toprule
\multirow{2}{*}{\textbf{Target Dimension}} &\multirow{2}{*}{\textbf{k1}} &\multirow{2}{*}{\textbf{k2}} &\multirow{2}{*}{\textbf{Stress}} &\multirow{2}{*}{\textbf{PCA Stress}} &\multicolumn{2}{c}{\textbf{RMap Stress ($\alpha$=20)}} &\multirow{2}{*}{\textbf{S-PCA Stress}} &\multirow{2}{*}{\textbf{K-PCA Stress}} &\multirow{2}{*}{\textbf{UMAP Stress}} \\\cmidrule{6-7}
& & & & &$\mathbf{\mu}$ &$\mathbf{\sigma}$ & & & \\\midrule
10 &0 &10 &0.150986 &\multirow{7}{*}{0.21} &\multirow{7}{*}{0.16} &\multirow{7}{*}{0.009} &\multirow{7}{*}{0.24} &\multirow{7}{*}{0.69} &\multirow{7}{*}{1.26} \\
10 &2 &8 &0.130287 & & & & & & \\
\textbf{10} &\textbf{3} &\textbf{7} &\textbf{0.127005} & & & & & & \\
10 &4 &6 &0.131711 & & & & & & \\
10 &5 &5 &0.134101 & & & & & & \\
10 &6 &4 &0.13699 & & & & & & \\
10 &7 &3 &0.149397 & & & & & & \\
\specialrule{0.3pt}{0.1pt}{0.1pt}
20 &0 &20 &0.10698 &\multirow{9}{*}{0.15} &\multirow{9}{*}{0.11} &\multirow{9}{*}{0.005} &\multirow{9}{*}{0.18} &\multirow{9}{*}{0.69} &\multirow{9}{*}{1.25} \\
20 &2 &18 &0.088584 & & & & & & \\
20 &4 &16 &0.080418 & & & & & & \\
20 &5 &15 &0.079099 & & & & & & \\
\textbf{20} &\textbf{8} &\textbf{12} &\textbf{0.076988} & & & & & & \\
20 &10 &10 &0.07744 & & & & & & \\
20 &12 &8 &0.080066 & & & & & & \\
20 &15 &5 &0.091193 & & & & & & \\
20 &18 &2 &0.125555 & & & & & & \\
\specialrule{0.3pt}{0.1pt}{0.1pt}
30 &0 &30 &0.087177 &\multirow{10}{*}{0.12} &\multirow{10}{*}{0.09} &\multirow{10}{*}{0.004} &\multirow{10}{*}{0.15} &\multirow{10}{*}{0.69} &\multirow{10}{*}{1.27} \\
30 &3 &27 &0.067666 & & & & & & \\
30 &5 &25 &0.060986 & & & & & & \\
30 &8 &22 &0.056502 & & & & & & \\
30 &12 &18 &0.054123 & & & & & & \\
\textbf{30} &\textbf{15} &\textbf{15} &\textbf{0.053645} & & & & & & \\
30 &18 &12 &0.055647 & & & & & & \\
30 &20 &10 &0.057995 & & & & & & \\
30 &25 &5 &0.073301 & & & & & & \\
30 &27 &3 &0.088451 & & & & & & \\
\specialrule{0.3pt}{0.1pt}{0.1pt}
40 &0 &40 &0.072628 &\multirow{11}{*}{0.11} &\multirow{11}{*}{0.08} &\multirow{11}{*}{0.003} &\multirow{11}{*}{0.13} &\multirow{11}{*}{0.69} &\multirow{11}{*}{1.27} \\
40 &4 &36 &0.053967 & & & & & & \\
40 &5 &35 &0.052083 & & & & & & \\
40 &8 &32 &0.046417 & & & & & & \\
40 &10 &30 &0.045487 & & & & & & \\
40 &15 &25 &0.042958 & & & & & & \\
40 &16 &24 &0.042068 & & & & & & \\
\textbf{40} &\textbf{20} &\textbf{20} &\textbf{0.041934} & & & & & & \\
40 &25 &15 &0.043438 & & & & & & \\
40 &30 &10 &0.047964 & & & & & & \\
40 &35 &5 &0.062596 & & & & & & \\
\bottomrule
\end{tabular}
\end{threeparttable}
}
\end{adjustwidth}

%% file: supp_tables/gene-dr-stress.tex
\begin{adjustwidth}{-2.5 cm}{-2.5 cm}\centering
\resizebox{0.9\columnwidth}{!}{
\begin{threeparttable}[p]
\caption{geneRNASeq: $\Lambda_S$}\label{tab:gene-dr-stress}
\scriptsize
\begin{tabular}{ccccccccccc}\toprule\toprule
\multirow{2}{*}{\textbf{Target Dimension}} &\multirow{2}{*}{\textbf{k1}} &\multirow{2}{*}{\textbf{k2}} &\multirow{2}{*}{\textbf{Stress}} &\multirow{2}{*}{\textbf{PCA Stress}} &\multicolumn{2}{c}{\textbf{RMap Stress ($\alpha$=20)}} &\multirow{2}{*}{\textbf{S-PCA Stress}} &\multirow{2}{*}{\textbf{K-PCA Stress}} &\multirow{2}{*}{\textbf{UMAP Stress}} \\\cmidrule{6-7}
& & & & &$\mathbf{\mu}$ &$\mathbf{\sigma}$ & & & \\\midrule
10 &0 &10 &0.154996 &\multirow{7}{*}{0.21} &\multirow{7}{*}{0.16} &\multirow{7}{*}{0.008} &\multirow{7}{*}{0.25} &\multirow{7}{*}{NA} &\multirow{7}{*}{18.72} \\
10 &2 &8 &0.156133 & & & & & & \\
10 &3 &7 &0.138719 & & & & & & \\
10 &4 &6 &0.132871 & & & & & & \\
\textbf{10} &\textbf{5} &\textbf{5} &\textbf{0.130126} & & & & & & \\
10 &6 &4 &0.133581 & & & & & & \\
10 &7 &3 &0.147557 & & & & & & \\
\specialrule{0.3pt}{0.1pt}{0.1pt}
20 &0 &20 &0.103773 &\multirow{10}{*}{0.17} &\multirow{10}{*}{0.11} &\multirow{10}{*}{0.006} &\multirow{10}{*}{0.24} &\multirow{10}{*}{0.70} &\multirow{10}{*}{18.60} \\
20 &2 &18 &0.091928 & & & & & & \\
20 &4 &16 &0.082728 & & & & & & \\
20 &5 &15 &0.080249 & & & & & & \\
20 &6 &14 &0.077367 & & & & & & \\
\textbf{20} &\textbf{8} &\textbf{12} &\textbf{0.071007} & & & & & & \\
20 &10 &10 &0.075024 & & & & & & \\
20 &12 &8 &0.078858 & & & & & & \\
20 &15 &5 &0.094134 & & & & & & \\
20 &18 &2 &0.134547 & & & & & & \\
\specialrule{0.3pt}{0.1pt}{0.1pt}
30 &0 &30 &0.092378 &\multirow{11}{*}{0.15} &\multirow{11}{*}{0.09} &\multirow{11}{*}{0.009} &\multirow{11}{*}{0.25} &\multirow{11}{*}{0.70} &\multirow{11}{*}{18.72} \\
30 &3 &27 &0.070738 & & & & & & \\
30 &5 &25 &0.059465 & & & & & & \\
30 &8 &22 &0.054121 & & & & & & \\
\textbf{30} &\textbf{10} &\textbf{20} &\textbf{0.052897} & & & & & & \\
30 &12 &18 &0.055643 & & & & & & \\
30 &15 &15 &0.057122 & & & & & & \\
30 &18 &12 &0.059765 & & & & & & \\
30 &20 &10 &0.062337 & & & & & & \\
30 &25 &5 &0.083422 & & & & & & \\
30 &27 &3 &0.101695 & & & & & & \\
\specialrule{0.3pt}{0.1pt}{0.1pt}
40 &0 &40 &0.090848 &\multirow{11}{*}{0.14} &\multirow{11}{*}{0.08} &\multirow{11}{*}{0.004} &\multirow{11}{*}{0.25} &\multirow{11}{*}{0.70} &\multirow{11}{*}{18.22} \\
40 &4 &36 &0.055797 & & & & & & \\
40 &5 &35 &0.050136 & & & & & & \\
40 &8 &32 &0.045299 & & & & & & \\
\textbf{40} &\textbf{10} &\textbf{30} &\textbf{0.043141} & & & & & & \\
40 &15 &25 &0.043549 & & & & & & \\
40 &16 &24 &0.044106 & & & & & & \\
40 &20 &20 &0.045514 & & & & & & \\
40 &25 &15 &0.049506 & & & & & & \\
40 &30 &10 &0.05636 & & & & & & \\
40 &35 &5 &0.077122 & & & & & & \\
\bottomrule
\end{tabular}
\end{threeparttable}
}
\end{adjustwidth}

%% file: supp_tables/reuters-dr-stress.tex
\squeezeup
\vspace*{-\baselineskip}
\begin{adjustwidth}{-2.5 cm}{-2.5 cm}\centering
\resizebox{\columnwidth}{!}{
\begin{threeparttable}[p]
\caption{Reuters30k: $\Lambda_S$}\label{tab:reuters-dr-stress}
\scriptsize
\begin{tabular}{ccccccccccc}\toprule\toprule
\multirow{2}{*}{\textbf{Target Dimension}} &\multirow{2}{*}{\textbf{k1}} &\multirow{2}{*}{\textbf{k2}} &\multirow{2}{*}{\textbf{Stress}} &\multirow{2}{*}{\textbf{PCA Stress}} &\multicolumn{2}{c}{\textbf{RMap Stress ($\alpha$=20)}} &\multirow{2}{*}{\textbf{S-PCA Stress}} &\multirow{2}{*}{\textbf{K-PCA Stress}} &\multirow{2}{*}{\textbf{UMAP Stress}} \\\cmidrule{6-7}
& & & & &$\mathbf{\mu}$ &$\mathbf{\sigma}$ & & & \\\midrule
\textbf{10} &\textbf{0} &\textbf{10} &\textbf{0.155841} &\multirow{7}{*}{0.49} &\multirow{7}{*}{0.16} &\multirow{7}{*}{0.001} &\multirow{7}{*}{0.49} &\multirow{7}{*}{0.71} &\multirow{7}{*}{3.35} \\
10 &2 &8 &0.162356 & & & & & & \\
10 &3 &7 &0.170477 & & & & & & \\
10 &4 &6 &0.183243 & & & & & & \\
10 &5 &5 &0.197407 & & & & & & \\
10 &6 &4 &0.216498 & & & & & & \\
10 &7 &3 &0.244457 & & & & & & \\
\specialrule{0.3pt}{0.1pt}{0.1pt}
20 &0 &20 &0.110339 &\multirow{9}{*}{0.45} &\multirow{9}{*}{0.11} &\multirow{9}{*}{0.001} &\multirow{9}{*}{0.46} &\multirow{9}{*}{0.70} &\multirow{9}{*}{3.27} \\
\textbf{20} &\textbf{2} &\textbf{18} &\textbf{0.109416} & & & & & & \\
20 &4 &16 &0.114293 & & & & & & \\
20 &5 &15 &0.11665 & & & & & & \\
20 &8 &12 &0.127054 & & & & & & \\
20 &10 &10 &0.136838 & & & & & & \\
20 &12 &8 &0.150428 & & & & & & \\
20 &15 &5 &0.184754 & & & & & & \\
20 &18 &2 &0.27384 & & & & & & \\
\specialrule{0.3pt}{0.1pt}{0.1pt}
30 &0 &30 &0.090478 &\multirow{12}{*}{0.43} &\multirow{12}{*}{0.09} &\multirow{12}{*}{0.001} &\multirow{12}{*}{0.44} &\multirow{12}{*}{0.70} &\multirow{12}{*}{3.21} \\
\textbf{30} &\textbf{2} &\textbf{28} &\textbf{0.088198} & & & & & & \\
30 &3 &27 &0.088585 & & & & & & \\
30 &5 &25 &0.089782 & & & & & & \\
30 &8 &22 &0.094184 & & & & & & \\
30 &10 &20 &0.097164 & & & & & & \\
30 &12 &18 &0.101738 & & & & & & \\
30 &15 &15 &0.109841 & & & & & & \\
30 &18 &12 &0.120367 & & & & & & \\
30 &20 &10 &0.130392 & & & & & & \\
30 &25 &5 &0.179204 & & & & & & \\
30 &27 &3 &0.225727 & & & & & & \\
\specialrule{0.3pt}{0.1pt}{0.1pt}
40 &0 &40 &0.079027 &\multirow{12}{*}{0.41} &\multirow{12}{*}{0.08} &\multirow{12}{*}{0.001} &\multirow{12}{*}{0.43} &\multirow{12}{*}{0.70} &\multirow{12}{*}{3.12} \\
40 &2 &38 &0.075794 & & & & & & \\
\textbf{\textcolor{red}{40}} &\textbf{\textcolor{red}{4}} &\textbf{\textcolor{red}{36}} &\textbf{\textcolor{red}{0.075186}} & & & & & & \\
40 &5 &35 &0.076406 & & & & & & \\
40 &8 &32 &0.077578 & & & & & & \\
40 &10 &30 &0.079557 & & & & & & \\
40 &15 &25 &0.085225 & & & & & & \\
40 &16 &24 &0.086591 & & & & & & \\
40 &20 &20 &0.093946 & & & & & & \\
40 &25 &15 &0.10561 & & & & & & \\
40 &30 &10 &0.127771 & & & & & & \\
40 &35 &5 &0.175076 & & & & & & \\
\bottomrule
\end{tabular}
\end{threeparttable}
}
\end{adjustwidth}

%% file: supp_tables/aptos-dr-stress.tex
\begin{adjustwidth}{-2.5 cm}{-2.5 cm}\centering
\resizebox{\columnwidth}{!}{
\begin{threeparttable}[p]
\caption{APTOS 2019: $\Lambda_{S}$}\label{tab:aptos-dr-stress}
\scriptsize
\begin{tabular}{cccccccc}\toprule\toprule
\multirow{2}{*}{\textbf{Target Dimension}} &\multirow{2}{*}{\textbf{k1}} &\multirow{2}{*}{\textbf{k2}} &\multirow{2}{*}{\textbf{Stress}} &\multirow{2}{*}{\textbf{PCA Stress}} &\multicolumn{2}{c}{\textbf{RMap Stress ($\alpha$=20)}} \\\cmidrule{6-7}
& & & & &$\mathbf{\mu}$ &$\mathbf{\sigma}$ \\\midrule
10 &0 &10 &0.179073 &\multirow{8}{*}{0.12} &\multirow{8}{*}{0.16} &\multirow{8}{*}{0.016} \\
10 &1 &9 &0.148839 & & & \\
10 &2 &8 &0.123974 & & & \\
10 &3 &7 &0.122061 & & & \\
10 &4 &6 &0.1052 & & & \\
10 &5 &5 &0.097674 & & & \\
10 &6 &4 &0.101736 & & & \\
\textbf{10} &\textbf{7} &\textbf{3} &\textbf{0.097584} & & & \\
\specialrule{0.3pt}{0.1pt}{0.1pt}
20 &0 &20 &0.095128 &\multirow{10}{*}{0.08} &\multirow{10}{*}{0.11} &\multirow{10}{*}{0.014} \\
20 &2 &18 &0.085237 & & & \\
20 &3 &17 &0.075839 & & & \\
20 &4 &16 &0.06693 & & & \\
20 &5 &15 &0.058859 & & & \\
\textbf{20} &\textbf{8} &\textbf{12} &\textbf{0.045718} & & & \\
20 &10 &10 &0.047303 & & & \\
20 &12 &8 &0.046251 & & & \\
20 &15 &5 &0.05158 & & & \\
20 &18 &2 &0.06905 & & & \\
\specialrule{0.3pt}{0.1pt}{0.1pt}
30 &0 &30 &0.091059 &\multirow{13}{*}{0.06} &\multirow{13}{*}{0.09} &\multirow{13}{*}{0.008} \\
30 &2 &28 &0.060186 & & & \\
30 &3 &27 &0.060056 & & & \\
30 &5 &25 &0.046503 & & & \\
30 &8 &22 &0.034993 & & & \\
30 &10 &20 &0.032974 & & & \\
30 &11 &19 &0.031663 & & & \\
30 &12 &18 &0.030824 & & & \\
\textbf{30} &\textbf{15} &\textbf{15} &\textbf{0.030116} & & & \\
30 &18 &12 &0.030801 & & & \\
30 &20 &10 &0.030715 & & & \\
30 &25 &5 &0.038002 & & & \\
30 &27 &3 &0.045107 & & & \\
\specialrule{0.3pt}{0.1pt}{0.1pt}
40 &0 &40 &0.086019 &\multirow{12}{*}{0.05} &\multirow{12}{*}{0.08} &\multirow{12}{*}{0.012} \\
40 &2 &38 &0.055444 & & & \\
40 &4 &36 &0.042809 & & & \\
40 &5 &35 &0.038921 & & & \\
40 &8 &32 &0.029978 & & & \\
40 &10 &30 &0.026612 & & & \\
40 &15 &25 &0.023963 & & & \\
40 &16 &24 &0.024174 & & & \\
\textbf{40} &\textbf{20} &\textbf{20} &\textbf{0.022142} & & & \\
40 &25 &15 &0.022693 & & & \\
40 &30 &10 &0.025446 & & & \\
40 &35 &5 &0.031453 & & & \\
\bottomrule
\end{tabular}
\end{threeparttable}
}
\end{adjustwidth}

%% file: supp_tables/div2k-dr-stress.tex
\begin{adjustwidth}{-2.5 cm}{-2.5 cm}\centering
\resizebox{0.9\columnwidth}{!}{
\begin{threeparttable}[p]
\caption{DIV2k: $\Lambda_S$}\label{tab:div2k-dr-stress}
\scriptsize
\begin{tabular}{cccccccc}\toprule\toprule
\multirow{2}{*}{\textbf{Target Dimension}} &\multirow{2}{*}{\textbf{k1}} &\multirow{2}{*}{\textbf{k2}} &\multirow{2}{*}{\textbf{Stress}} &\multirow{2}{*}{\textbf{PCA Stress}} &\multicolumn{2}{c}{\textbf{RMap Stress ($\alpha$=20)}} \\\cmidrule{6-7}
& & & & &$\mathbf{\mu}$ &$\mathbf{\sigma}$ \\\midrule
10 &0 &10 &0.156703 &\multirow{8}{*}{0.31} &\multirow{8}{*}{0.16} &\multirow{8}{*}{0.003} \\
\textbf{10} &\textbf{1} &\textbf{9} &\textbf{0.144361} & & & \\
10 &2 &8 &0.144838 & & & \\
10 &3 &7 &0.146517 & & & \\
10 &4 &6 &0.149547 & & & \\
10 &5 &5 &0.157161 & & & \\
10 &6 &4 &0.174424 & & & \\
10 &7 &3 &0.185904 & & & \\
\specialrule{0.3pt}{0.1pt}{0.1pt}
20 &0 &20 &0.109547 &\multirow{10}{*}{0.26} &\multirow{10}{*}{0.11} &\multirow{10}{*}{0.003} \\
20 &2 &18 &0.099879 & & & \\
20 &3 &17 &0.09588 & & & \\
20 &4 &16 &0.093435 & & & \\
\textbf{20} &\textbf{5} &\textbf{15} &\textbf{0.091908} & & & \\
20 &8 &12 &0.096338 & & & \\
20 &10 &10 &0.102482 & & & \\
20 &12 &8 &0.109068 & & & \\
20 &15 &5 &0.132754 & & & \\
20 &18 &2 &0.190162 & & & \\
\specialrule{0.3pt}{0.1pt}{0.1pt}
30 &0 &30 &0.09143 &\multirow{12}{*}{0.23} &\multirow{12}{*}{0.09} &\multirow{12}{*}{0.002} \\
30 &2 &28 &0.081431 & & & \\
30 &3 &27 &0.076753 & & & \\
30 &5 &25 &0.072565 & & & \\
30 &8 &22 &0.073153 & & & \\
\textbf{30} &\textbf{10} &\textbf{20} &\textbf{0.072457} & & & \\
30 &12 &18 &0.073059 & & & \\
30 &15 &15 &0.078949 & & & \\
30 &18 &12 &0.083008 & & & \\
30 &20 &10 &0.090443 & & & \\
30 &25 &5 &0.121266 & & & \\
30 &27 &3 &0.147061 & & & \\
\specialrule{0.3pt}{0.1pt}{0.1pt}
40 &0 &40 &0.080119 &\multirow{12}{*}{0.21} &\multirow{12}{*}{0.08} &\multirow{12}{*}{0.002} \\
40 &2 &38 &0.069251 & & & \\
40 &4 &36 &0.064546 & & & \\
40 &5 &35 &0.061345 & & & \\
40 &8 &32 &0.061543 & & & \\
\textbf{40} &\textbf{10} &\textbf{30} &\textbf{0.05954} & & & \\
40 &15 &25 &0.062452 & & & \\
40 &16 &24 &0.061158 & & & \\
40 &20 &20 &0.064559 & & & \\
40 &25 &15 &0.06917 & & & \\
40 &30 &10 &0.083609 & & & \\
40 &35 &5 &0.109629 & & & \\
\bottomrule
\end{tabular}
\end{threeparttable}
}
\end{adjustwidth}

%% file: supp_tables/bank-dr-m1.tex
\begin{adjustwidth}{-2.5 cm}{-2.5 cm}\centering
\resizebox{0.88\columnwidth}{!}{
\begin{threeparttable}[p]
\caption{Bank: $\Lambda_{M_1}$}\label{tab:bank-dr-m1}
\scriptsize
\begin{tabular}{ccccccccccc}\toprule\toprule
\multirow{2}{*}{\textbf{Target Dimension}} &\multirow{2}{*}{\textbf{k1}} 
&\multirow{2}{*}{\textbf{k2}} &\multirow{2}{*}{\textbf{M1}} &\multirow{2}{*}
{\textbf{PCA M1}} &\multicolumn{2}{c}{\textbf{RMap M1 ($\mathbf{\alpha}$=20)}} &\multirow{2}{*}{\textbf{S-PCA M1}} &\multirow{2}{*}{\textbf{K-PCA M1}} &\multirow{2}{*}{\textbf{UMAP M1}} \\
& & & & &$\mathbf{\mu}$ &$\mathbf{\sigma}$ & & & \\\midrule
1 &0 &1 &0.006966 &0.651017 &0.61 &0.425 &0.683462 &0.960012 &171.177790 \\
\specialrule{0.3pt}{0.1pt}{0.1pt}
2 &0 &2 &0.012508 &\multirow{2}{*}{0.584043} &\multirow{2}{*}{0.42} &\multirow{2}{*}{0.229} &\multirow{2}{*}{0.614852} &\multirow{2}{*}{0.951856} &\multirow{2}{*}{95.028671} \\
\textbf{2} &\textbf{1} &\textbf{1} &\textbf{0.012201} & & & & & & \\
\specialrule{0.3pt}{0.1pt}{0.1pt}
3 &0 &3 &0.001685 &\multirow{3}{*}{0.550702} &\multirow{3}{*}{0.43} &\multirow{3}{*}{0.265} &\multirow{3}{*}{0.577230} &\multirow{3}{*}{0.947818} &\multirow{3}{*}{94.888941} \\
3 &1 &2 &0.003476 & & & & & & \\
\textbf{3} &\textbf{2} &\textbf{1} &\textbf{0.000357} & & & & & & \\
\specialrule{0.3pt}{0.1pt}{0.1pt}
5 &0 &5 &0.002059 &\multirow{5}{*}{0.535537} &\multirow{5}{*}{0.38} &\multirow{5}{*}{0.446} &\multirow{5}{*}{0.564456} &\multirow{5}{*}{0.945817} &\multirow{5}{*}{122.448199} \\
5 &1 &4 &2.51e-5 & & & & & & \\
5 &2 &3 &2.82e-5 & & & & & & \\
5 &3 &2 &0.000148 & & & & & & \\
\textbf{5} &\textbf{4} &\textbf{1} &\textbf{5.82e-6} & & & & & & \\
\specialrule{0.3pt}{0.1pt}{0.1pt}
6 &0 &6 &0.002623 &\multirow{6}{*}{0.535173} &\multirow{6}{*}{0.23} &\multirow{6}{*}{0.204} &\multirow{6}{*}{0.576306} &\multirow{6}{*}{0.945657} &\multirow{6}{*}{182.743756} \\
6 &1 &5 &0.004556 & & & & & & \\
6 &2 &4 &3.60e-5 & & & & & & \\
6 &3 &3 &0.000192 & & & & & & \\
6 &4 &2 &2.92e-5 & & & & & & \\
\textbf{6} &\textbf{5} &\textbf{1} &\textbf{4.57e-7} & & & & & & \\
\specialrule{0.3pt}{0.1pt}{0.1pt}
7 &0 &7 &0.011784 &\multirow{7}{*}{0.534967} &\multirow{7}{*}{0.26} &\multirow{7}{*}{0.234} &\multirow{7}{*}{0.575984} &\multirow{7}{*}{0.945562} &\multirow{7}{*}{190.630670} \\
7 &1 &6 &0.003607 & & & & & & \\
7 &2 &5 &1.94e-5 & & & & & & \\
7 &3 &4 &0.000287 & & & & & & \\
7 &4 &3 &3.31e-6 & & & & & & \\
\textbf{7} &\textbf{5} &\textbf{2} &\textbf{2.59e-6} & & & & & & \\
7 &6 &1 &9.22e-6 & & & & & & \\
\specialrule{0.3pt}{0.1pt}{0.1pt}
8 &0 &8 &0.001774 &\multirow{7}{*}{0.534825} &\multirow{7}{*}{0.23} &\multirow{7}{*}{0.202} &\multirow{7}{*}{0.575932} &\multirow{7}{*}{0.945489} &\multirow{7}{*}{256.766924} \\
8 &2 &6 &0.000117 & & & & & & \\
8 &3 &5 &2.54e-5 & & & & & & \\
8 &4 &4 &1.01e-5 & & & & & & \\
8 &5 &3 &7.06e-6 & & & & & & \\
8 &6 &2 &3.57e-6 & & & & & & \\
\textbf{8} &\textbf{7} &\textbf{1} &\textbf{3.49e-7} & & & & & & \\
\bottomrule
\end{tabular}
\end{threeparttable}
}
\end{adjustwidth}
\vfill

%% file: supp_tables/hatespeech-dr-m1.tex
\begin{adjustwidth}{-2.5 cm}{-2.5 cm}\centering
\resizebox{0.88\columnwidth}{!}{
\begin{threeparttable}[p]
\caption{Hatespeech: $\Lambda_{M_1}$}\label{tab:hatespeech-dr-m1}
\scriptsize
\begin{tabular}{ccccccccccc}\toprule\toprule
\multirow{2}{*}{\textbf{Target Dimension}} &\multirow{2}{*}{\textbf{k1}} &\multirow{2}{*}{\textbf{k2}} &\multirow{2}{*}{\textbf{M1}} &\multirow{2}{*}{\textbf{PCA M1}} &\multicolumn{2}{c}{\textbf{RMap M1 ($\mathbf{\alpha}$=20)}} &\multirow{2}{*}{\textbf{S-PCA M1}} &\multirow{2}{*}{\textbf{K-PCA M1}} &\multirow{2}{*}{\textbf{UMAP M1}} \\\cmidrule{6-7}
& & & & &$\mathbf{\mu}$ &$\mathbf{\sigma}$ & & & \\\midrule
10 &0 &10 &0.001827 &\multirow{8}{*}{0.66} &\multirow{8}{*}{0.06} &\multirow{8}{*}{0.04} &\multirow{8}{*}{0.68} &\multirow{8}{*}{0.99} &\multirow{8}{*}{240.50} \\
10 &1 &9 &0.001306 & & & & & & \\
10 &2 &8 &4.47e-4 & & & & & & \\
10 &3 &7 &0.001135 & & & & & & \\
10 &4 &6 &0.00197 & & & & & & \\
10 &5 &5 &0.002305 & & & & & & \\
\textbf{10} &\textbf{6} &\textbf{4} &\textbf{0.000191} & & & & & & \\
10 &7 &3 &0.000364 & & & & & & \\
\specialrule{0.3pt}{0.1pt}{0.1pt}
20 &0 &20 &1.64e-3 &\multirow{10}{*}{0.50} &\multirow{10}{*}{0.04} &\multirow{10}{*}{0.03} &\multirow{10}{*}{0.53} &\multirow{10}{*}{0.99} &\multirow{10}{*}{569.90} \\
20 &2 &18 &0.000502 & & & & & & \\
20 &3 &17 &1.36e-3 & & & & & & \\
\textbf{20} &\textbf{4} &\textbf{16} &\textbf{9.59e-6} & & & & & & \\
20 &5 &15 &0.000354 & & & & & & \\
20 &8 &12 &6.57e-4 & & & & & & \\
20 &10 &10 &0.000718 & & & & & & \\
20 &12 &8 &0.000512 & & & & & & \\
20 &15 &5 &0.000965 & & & & & & \\
20 &18 &2 &0.000668 & & & & & & \\
\specialrule{0.3pt}{0.1pt}{0.1pt}
30 &0 &30 &0.00033 &\multirow{13}{*}{0.41} &\multirow{13}{*}{0.03} &\multirow{13}{*}{0.02} &\multirow{13}{*}{0.44} &\multirow{13}{*}{0.99} &\multirow{13}{*}{828.16} \\
\textbf{30} &\textbf{2} &\textbf{28} &\textbf{1.76e-6} & & & & & & \\
30 &3 &27 &0.000218 & & & & & & \\
30 &5 &25 &2.21e-3 & & & & & & \\
30 &6 &24 &3.64e-4 & & & & & & \\
30 &8 &22 &6.39e-4 & & & & & & \\
30 &10 &20 &1.82e-4 & & & & & & \\
30 &12 &18 &4.03e-4 & & & & & & \\
30 &15 &15 &5.34e-5 & & & & & & \\
30 &18 &12 &1.54e-4 & & & & & & \\
30 &20 &10 &3.40e-6 & & & & & & \\
30 &25 &5 &7.39e-5 & & & & & & \\
30 &27 &3 &3.76e-4 & & & & & & \\
\specialrule{0.3pt}{0.1pt}{0.1pt}
40 &0 &40 &1.26e-3 &\multirow{13}{*}{0.33} &\multirow{13}{*}{0.03} &\multirow{13}{*}{0.02} &\multirow{13}{*}{0.36} &\multirow{13}{*}{0.99} &\multirow{13}{*}{1095.06} \\
40 &2 &38 &0.00123 & & & & & & \\
40 &4 &36 &3.35e-4 & & & & & & \\
40 &5 &35 &2.42e-4 & & & & & & \\
40 &8 &32 &7.41e-5 & & & & & & \\
40 &10 &30 &3.36e-5 & & & & & & \\
40 &11 &29 &2.31e-4 & & & & & & \\
40 &15 &25 &2.14e-4 & & & & & & \\
40 &16 &24 &0.000412 & & & & & & \\
\textbf{40} &\textbf{20} &\textbf{20} &\textbf{5.38e-6} & & & & & & \\
40 &25 &15 &0.000328 & & & & & & \\
40 &30 &10 &0.000132 & & & & & & \\
40 &35 &5 &8.17e-5 & & & & & & \\
\bottomrule
\end{tabular}
\end{threeparttable}
}
\end{adjustwidth}

%% file: supp_tables/fmnist-dr-m1.tex
\begin{adjustwidth}{-2.5 cm}{-2.5 cm}\centering
\resizebox{0.88\columnwidth}{!}{
\begin{threeparttable}[p]
\caption{FMnist: $\Lambda_{M_1}$}\label{tab:fmnist-dr-m1}
\scriptsize
\begin{tabular}{ccccccccccc}\toprule\toprule
\multirow{2}{*}{\textbf{Target Dimension}} &\multirow{2}{*}{\textbf{k1}} &\multirow{2}{*}{\textbf{k2}} &\multirow{2}{*}{\textbf{M1}} &\multirow{2}{*}{\textbf{PCA M1}} &\multicolumn{2}{c}{\textbf{RMap M1 ($\mathbf{\alpha}$=20)}} &\multirow{2}{*}{\textbf{S-PCA M1}} &\multirow{2}{*}{\textbf{K-PCA M1}} &\multirow{2}{*}{\textbf{UMAP M1}} \\\cmidrule{6-7}
& & & & &$\mathbf{\mu}$ &$\mathbf{\sigma}$ & & & \\\midrule
10 &0 &10 &0.005264 &\multirow{7}{*}{0.60} &\multirow{7}{*}{0.11} &\multirow{7}{*}{0.059} &\multirow{7}{*}{0.64} &\multirow{7}{*}{1.00} &\multirow{7}{*}{241.35} \\
10 &2 &8 &0.000153 & & & & & & \\
\textbf{10} &\textbf{3} &\textbf{7} &\textbf{3.74e-5} & & & & & & \\
10 &4 &6 &0.000303 & & & & & & \\
10 &5 &5 &0.000181 & & & & & & \\
10 &6 &4 &0.000192 & & & & & & \\
10 &7 &3 &0.000181 & & & & & & \\
\specialrule{0.3pt}{0.1pt}{0.1pt}
20 &0 &20 &0.001009 &\multirow{10}{*}{0.55} &\multirow{10}{*}{0.11} &\multirow{10}{*}{0.077} &\multirow{10}{*}{0.59} &\multirow{10}{*}{1.00} &\multirow{10}{*}{487.89} \\
20 &2 &18 &4.94e-5 & & & & & & \\
20 &4 &16 &0.000164 & & & & & & \\
20 &5 &15 &3.21e-5 & & & & & & \\
\textbf{20} &\textbf{6} &\textbf{14} &\textbf{1.79e-5} & & & & & & \\
20 &8 &12 &0.000104 & & & & & & \\
20 &10 &10 &3.71e-5 & & & & & & \\
20 &12 &8 &0.000122 & & & & & & \\
20 &15 &5 &0.000289 & & & & & & \\
20 &18 &2 &0.000163 & & & & & & \\
\specialrule{0.3pt}{0.1pt}{0.1pt}
30 &0 &30 &0.001011 &\multirow{11}{*}{0.52} &\multirow{11}{*}{0.09} &\multirow{11}{*}{0.054} &\multirow{11}{*}{0.56} &\multirow{11}{*}{1.00} &\multirow{11}{*}{781.17} \\
30 &3 &27 &0.00012 & & & & & & \\
30 &5 &25 &4.63e-5 & & & & & & \\
30 &8 &22 &0.000116 & & & & & & \\
30 &10 &20 &2.21e-5 & & & & & & \\
30 &12 &18 &5.55e-5 & & & & & & \\
30 &15 &15 &2.46e-5 & & & & & & \\
30 &18 &12 &4.09e-5 & & & & & & \\
30 &20 &10 &9.53e-6 & & & & & & \\
\textbf{30} &\textbf{25} &\textbf{5} &\textbf{2.73e-7} & & & & & & \\
30 &27 &3 &9.15e-6 & & & & & & \\
\specialrule{0.3pt}{0.1pt}{0.1pt}
40 &0 &40 &0.000826 &\multirow{11}{*}{0.49} &\multirow{11}{*}{0.09} &\multirow{11}{*}{0.061} &\multirow{11}{*}{0.54} &\multirow{11}{*}{1.00} &\multirow{11}{*}{1030.35} \\
40 &4 &36 &0.000212 & & & & & & \\
40 &5 &35 &2.01e-5 & & & & & & \\
40 &8 &32 &5.20e-5 & & & & & & \\
40 &10 &30 &0.000129 & & & & & & \\
\textbf{40} &\textbf{15} &\textbf{25} &\textbf{1.10e-6} & & & & & & \\
40 &16 &24 &1.68e-6 & & & & & & \\
40 &20 &20 &2.02e-5 & & & & & & \\
40 &25 &15 &2.39e-5 & & & & & & \\
40 &30 &10 &3.05e-6 & & & & & & \\
40 &35 &5 &2.16e-5 & & & & & & \\
\bottomrule
\end{tabular}
\end{threeparttable}
}
\end{adjustwidth}
\vfill

%% file: supp_tables/cifar-dr-m1.tex
\begin{adjustwidth}{-2.5 cm}{-2.5 cm}\centering
\resizebox{0.88\columnwidth}{!}{
\begin{threeparttable}[p]
\caption{Cifar10: $\Lambda_{M_1}$}\label{tab:cifar-dr-m1}
\scriptsize
\begin{tabular}{ccccccccccc}\toprule\toprule
\multirow{2}{*}{\textbf{Target Dimension}} &\multirow{2}{*}{\textbf{k1}} &\multirow{2}{*}{\textbf{k2}} &\multirow{2}{*}{\textbf{M1}} &\multirow{2}{*}{\textbf{PCA M1}} &\multicolumn{2}{c}{\textbf{RMap M1 ($\mathbf{\alpha}$=20)}} &\multirow{2}{*}{\textbf{S-PCA M1}} &\multirow{2}{*}{\textbf{K-PCA M1}} &\multirow{2}{*}{\textbf{UMAP M1}} \\\cmidrule{6-7}
& & & & &$\mathbf{\mu}$ &$\mathbf{\sigma}$ & & & \\\midrule
10 &0 &10 &0.002759 &\multirow{7}{*}{0.49} &\multirow{7}{*}{0.09} &\multirow{7}{*}{0.062} &\multirow{7}{*}{0.54} &\multirow{7}{*}{1.00} &\multirow{7}{*}{166.84} \\
10 &2 &8 &0.000142 & & & & & & \\
10 &3 &7 &0.000154 & & & & & & \\
10 &4 &6 &0.000156 & & & & & & \\
10 &5 &5 &0.001149 & & & & & & \\
\textbf{10} &\textbf{6} &\textbf{4} &\textbf{0.000131} & & & & & & \\
\specialrule{0.3pt}{0.1pt}{0.1pt}
10 &7 &3 &0.000799 & & & & & & \\
20 &0 &20 &0.000342 &\multirow{9}{*}{0.38} &\multirow{9}{*}{0.04} &\multirow{9}{*}{0.035} &\multirow{9}{*}{0.43} &\multirow{9}{*}{1.00} &\multirow{9}{*}{485.22} \\
20 &2 &18 &0.000517 & & & & & & \\
20 &4 &16 &0.000547 & & & & & & \\
20 &5 &15 &0.001161 & & & & & & \\
\textbf{20} &\textbf{8} &\textbf{12} &\textbf{0.00011} & & & & & & \\
20 &10 &10 &0.000162 & & & & & & \\
20 &12 &8 &0.000167 & & & & & & \\
20 &15 &5 &0.000416 & & & & & & \\
20 &18 &2 &0.000236 & & & & & & \\
\specialrule{0.3pt}{0.1pt}{0.1pt}
30 &0 &30 &0.001148 &\multirow{10}{*}{0.32} &\multirow{10}{*}{0.05} &\multirow{10}{*}{0.031} &\multirow{10}{*}{0.38} &\multirow{10}{*}{1.00} &\multirow{10}{*}{753.84} \\
30 &3 &27 &0.000331 & & & & & & \\
30 &5 &25 &0.000243 & & & & & & \\
30 &8 &22 &0.000126 & & & & & & \\
30 &12 &18 &0.000454 & & & & & & \\
\textbf{30} &\textbf{15} &\textbf{15} &\textbf{5.66e-5} & & & & & & \\
30 &18 &12 &0.00086 & & & & & & \\
30 &20 &10 &6.23e-5 & & & & & & \\
30 &25 &5 &9.57e-5 & & & & & & \\
30 &27 &3 &0.000344 & & & & & & \\
\specialrule{0.3pt}{0.1pt}{0.1pt}
40 &0 &40 &0.000331 &\multirow{11}{*}{0.28} &\multirow{11}{*}{0.04} &\multirow{11}{*}{0.028} &\multirow{11}{*}{0.34} &\multirow{11}{*}{1.00} &\multirow{11}{*}{1008.78} \\
40 &4 &36 &0.001214 & & & & & & \\
40 &5 &35 &0.000229 & & & & & & \\
40 &8 &32 &0.000165 & & & & & & \\
40 &10 &30 &2.01e-5 & & & & & & \\
40 &15 &25 &2.84e-5 & & & & & & \\
40 &16 &24 &2.03e-4 & & & & & & \\
40 &20 &20 &4.94e-5 & & & & & & \\
40 &25 &15 &0.000398 & & & & & & \\
40 &30 &10 &1.12e-4 & & & & & & \\
\textbf{40} &\textbf{35} &\textbf{5} &\textbf{1.35e-5} & & & & & & \\
\bottomrule
\end{tabular}
\end{threeparttable}
}
\end{adjustwidth}

%% file: supp_tables/gene-dr-m1.tex
\begin{adjustwidth}{-2.5 cm}{-2.5 cm}\centering
\resizebox{0.87\columnwidth}{!}{
\begin{threeparttable}[p]
\caption{geneRNASeq: $\Lambda_{M_1}$}\label{tab:gene-dr-m1}
\scriptsize
\begin{tabular}{ccccccccccc}\toprule\toprule
\multirow{2}{*}{\textbf{Target Dimension}} &\multirow{2}{*}{\textbf{k1}} &\multirow{2}{*}{\textbf{k2}} &\multirow{2}{*}{\textbf{M1}} &\multirow{2}{*}{\textbf{PCA M1}} &\multicolumn{2}{c}{\textbf{RMap M1 ($\mathbf{\alpha}$=20)}} &\multirow{2}{*}{\textbf{S-PCA M1}} &\multirow{2}{*}{\textbf{K-PCA M1}} &\multirow{2}{*}{\textbf{UMAP M1}} \\\cmidrule{6-7}
& & & & &$\mathbf{\mu}$ &$\mathbf{\sigma}$ & & & \\\midrule
10 &0 &10 &0.009101 &\multirow{7}{*}{0.94} &\multirow{7}{*}{0.31} &\multirow{7}{*}{0.246} &\multirow{7}{*}{0.95} &\multirow{7}{*}{1.00} &\multirow{7}{*}{328.72} \\
10 &2 &8 &0.000113 & & & & & & \\
10 &3 &7 &0.000175 & & & & & & \\
10 &4 &6 &3.68e-5 & & & & & & \\
\textbf{10} &\textbf{5} &\textbf{5} &\textbf{1.69e-5} & & & & & & \\
10 &6 &4 &7.96e-5 & & & & & & \\
10 &7 &3 &6.20e-5 & & & & & & \\
\specialrule{0.3pt}{0.1pt}{0.1pt}
20 &0 &20 &0.004859 &\multirow{10}{*}{0.93} &\multirow{10}{*}{0.18} &\multirow{10}{*}{0.149} &\multirow{10}{*}{0.95} &\multirow{10}{*}{1.00} &\multirow{10}{*}{586.64} \\
20 &2 &18 &8.93e-5 & & & & & & \\
20 &4 &16 &0.000114 & & & & & & \\
\textbf{20} &\textbf{5} &\textbf{15} &\textbf{2.59e-6} & & & & & & \\
20 &6 &14 &8.39e-6 & & & & & & \\
20 &8 &12 &2.39e-5 & & & & & & \\
20 &10 &10 &1.58e-5 & & & & & & \\
20 &12 &8 &4.68e-5 & & & & & & \\
20 &15 &5 &8.67e-6 & & & & & & \\
20 &18 &2 &0.000121 & & & & & & \\
\specialrule{0.3pt}{0.1pt}{0.1pt}
30 &0 &30 &0.001417 &\multirow{11}{*}{0.93} &\multirow{11}{*}{0.19} &\multirow{11}{*}{0.143} &\multirow{11}{*}{0.95} &\multirow{11}{*}{1.00} &\multirow{11}{*}{826.54} \\
\textbf{30} &\textbf{3} &\textbf{27} &\textbf{1.25e-6} & & & & & & \\
30 &5 &25 &2.02e-5 & & & & & & \\
30 &8 &22 &1.51e-5 & & & & & & \\
30 &10 &20 &6.49e-6 & & & & & & \\
30 &12 &18 &1.53e-6 & & & & & & \\
30 &15 &15 &5.31e-6 & & & & & & \\
30 &18 &12 &2.33e-5 & & & & & & \\
30 &20 &10 &6.10e-6 & & & & & & \\
30 &25 &5 &2.02e-6 & & & & & & \\
30 &27 &3 &8.15e-6 & & & & & & \\
\specialrule{0.3pt}{0.1pt}{0.1pt}
40 &0 &40 &0.001658 &\multirow{11}{*}{0.93} &\multirow{11}{*}{0.12} &\multirow{11}{*}{0.081} &\multirow{11}{*}{0.95} &\multirow{11}{*}{1.00} &\multirow{11}{*}{1089.52} \\
40 &4 &36 &6.77e-6 & & & & & & \\
40 &5 &35 &1.25e-5 & & & & & & \\
40 &8 &32 &1.02e-5 & & & & & & \\
40 &10 &30 &1.38e-5 & & & & & & \\
\textbf{40} &\textbf{15} &\textbf{25} &\textbf{4.61e-6} & & & & & & \\
40 &16 &24 &1.61e-5 & & & & & & \\
40 &20 &20 &1.07e-5 & & & & & & \\
40 &25 &15 &2.02e-5 & & & & & & \\
40 &30 &10 &4.68e-6 & & & & & & \\
40 &35 &5 &1.48e-5 & & & & & & \\
\bottomrule
\end{tabular}
\end{threeparttable}
}
\end{adjustwidth}
\vfill

%% file: supp_tables/reuters-dr-m1.tex
\begin{adjustwidth}{-2.5 cm}{-2.5 cm}\centering
\resizebox{0.87\columnwidth}{!}{
\begin{threeparttable}[p]
\caption{Reuters30k: $\Lambda_{M_1}$}\label{tab:reuters-dr-m1}
\scriptsize
\begin{tabular}{ccccccccccc}\toprule\toprule
\multirow{2}{*}{\textbf{Target Dimension}} &\multirow{2}{*}{\textbf{k1}} &\multirow{2}{*}{\textbf{k2}} &\multirow{2}{*}{\textbf{M1}} &\multirow{2}{*}{\textbf{PCA M1}} &\multicolumn{2}{c}{\textbf{RMap M1 ($\mathbf{\alpha}$=20)}} &\multirow{2}{*}{\textbf{S-PCA M1}} &\multirow{2}{*}{\textbf{K-PCA M1}} &\multirow{2}{*}{\textbf{UMAP M1}} \\\cmidrule{6-7}
& & & & &$\mathbf{\mu}$ &$\mathbf{\sigma}$ & & & \\\midrule
10 &0 &10 &0.00062 &\multirow{7}{*}{0.88} &\multirow{7}{*}{0.03} &\multirow{7}{*}{0.018} &\multirow{7}{*}{0.88} &\multirow{7}{*}{1.00} &\multirow{7}{*}{196.97} \\
10 &2 &8 &7.17e-5 & & & & & & \\
10 &3 &7 &0.000112 & & & & & & \\
10 &4 &6 &0.000445 & & & & & & \\
\textbf{10} &\textbf{5} &\textbf{5} &\textbf{9.61e-6} & & & & & & \\
10 &6 &4 &0.000127 & & & & & & \\
10 &7 &3 &6.91e-5 & & & & & & \\
\specialrule{0.3pt}{0.1pt}{0.1pt}
20 &0 &20 &0.000577 &\multirow{9}{*}{0.85} &\multirow{9}{*}{0.03} &\multirow{9}{*}{0.020} &\multirow{9}{*}{0.85} &\multirow{9}{*}{1.00} &\multirow{9}{*}{394.25} \\
20 &2 &18 &2.30e-5 & & & & & & \\
20 &4 &16 &6.20e-5 & & & & & & \\
20 &5 &15 &0.000112 & & & & & & \\
20 &8 &12 &2.88e-5 & & & & & & \\
20 &10 &10 &8.41e-5 & & & & & & \\
\textbf{20} &\textbf{12} &\textbf{8} &\textbf{2.10e-5} & & & & & & \\
20 &15 &5 &0.000129 & & & & & & \\
20 &18 &2 &0.000133 & & & & & & \\
\specialrule{0.3pt}{0.1pt}{0.1pt}
30 &0 &30 &1.83e-5 &\multirow{12}{*}{0.83} &\multirow{12}{*}{0.02} &\multirow{12}{*}{0.017} &\multirow{12}{*}{0.84} &\multirow{12}{*}{1.00} &\multirow{12}{*}{679.72} \\
30 &2 &28 &NA & & & & & & \\
\textbf{30} &\textbf{3} &\textbf{27} &\textbf{1.50e-6} & & & & & & \\
30 &5 &25 &1.20e-4 & & & & & & \\
30 &8 &22 &3.25e-5 & & & & & & \\
30 &10 &20 &1.99e-5 & & & & & & \\
30 &12 &18 &2.53e-5 & & & & & & \\
30 &15 &15 &7.05e-5 & & & & & & \\
30 &18 &12 &1.60e-5 & & & & & & \\
30 &20 &10 &7.19e-5 & & & & & & \\
30 &25 &5 &7.04e-5 & & & & & & \\
30 &27 &3 &2.16e-5 & & & & & & \\
\specialrule{0.3pt}{0.1pt}{0.1pt}
40 &0 &40 &4.49e-5 &\multirow{12}{*}{0.81} &\multirow{12}{*}{0.02} &\multirow{12}{*}{0.014} &\multirow{12}{*}{0.83} &\multirow{12}{*}{1.00} &\multirow{12}{*}{913.86} \\
40 &2 &38 &NA & & & & & & \\
40 &4 &36 &1.72e-4 & & & & & & \\
40 &5 &35 &8.02e-5 & & & & & & \\
40 &8 &32 &4.53e-6 & & & & & & \\
40 &10 &30 &5.61e-5 & & & & & & \\
\textbf{40} &\textbf{15} &\textbf{25} &\textbf{5.17e-7} & & & & & & \\
40 &16 &24 &1.04e-5 & & & & & & \\
40 &20 &20 &4.85e-5 & & & & & & \\
40 &25 &15 &9.50e-6 & & & & & & \\
40 &30 &10 &4.60e-5 & & & & & & \\
40 &35 &5 &7.85e-5 & & & & & & \\
\bottomrule
\end{tabular}
\end{threeparttable}
}
\end{adjustwidth}

%% file: supp_tables/aptos-dr-m1.tex
\begin{adjustwidth}{-2.5 cm}{-2.5 cm}\centering
\resizebox{\columnwidth}{!}{
\begin{threeparttable}[p]
\caption{APTOS 2019: $\Lambda_{M_1}$}\label{tab:aptos-dr-m1}
\scriptsize
\begin{tabular}{cccccccc}\toprule\toprule
\multirow{2}{*}{\textbf{Target Dimension}} &\multirow{2}{*}{\textbf{k1}} &\multirow{2}{*}{\textbf{k2}} &\multirow{2}{*}{\textbf{M1}} &\multirow{2}{*}{\textbf{PCA M1}} &\multicolumn{2}{c}{\textbf{RMap M1 ($\mathbf{\alpha}$=20)}} \\\cmidrule{6-7}
& & & & &$\mathbf{\mu}$ &$\mathbf{\sigma}$ \\\midrule
10 &0 &10 &0.006698 &\multirow{8}{*}{0.81} &\multirow{8}{*}{0.24} &\multirow{8}{*}{0.15} \\
10 &1 &9 &0.000345 & & & \\
10 &2 &8 &0.000173 & & & \\
10 &3 &7 &2.47e-5 & & & \\
10 &4 &6 &2.98e-5 & & & \\
10 &5 &5 &3.82e-5 & & & \\
10 &6 &4 &4.09e-5 & & & \\
\textbf{10} &\textbf{7} &\textbf{3} &\textbf{1.58e-5} & & & \\
\specialrule{0.3pt}{0.1pt}{0.1pt}
20 &0 &20 &1.95e-3 &\multirow{10}{*}{0.79} &\multirow{10}{*}{0.15} &\multirow{10}{*}{0.12} \\
20 &2 &18 &8.24e-5 & & & \\
20 &3 &17 &1.27e-5 & & & \\
20 &4 &16 &5.36e-5 & & & \\
20 &5 &15 &2.09e-5 & & & \\
20 &8 &12 &2.58e-5 & & & \\
\textbf{20} &\textbf{10} &\textbf{10} &\textbf{2.76e-6} & & & \\
20 &12 &8 &4.15e-5 & & & \\
20 &15 &5 &7.27e-5 & & & \\
20 &18 &2 &8.11e-5 & & & \\
\specialrule{0.3pt}{0.1pt}{0.1pt}
30 &0 &30 &0.008812 &\multirow{13}{*}{0.78} &\multirow{13}{*}{0.15} &\multirow{13}{*}{0.08} \\
30 &2 &28 &6.57e-5 & & & \\
30 &3 &27 &9.98e-5 & & & \\
30 &5 &25 &9.81e-5 & & & \\
30 &8 &22 &3.99e-5 & & & \\
30 &10 &20 &4.82e-5 & & & \\
30 &11 &19 &1.57e-5 & & & \\
30 &12 &18 &9.66e-5 & & & \\
30 &15 &15 &1.87e-5 & & & \\
30 &18 &12 &1.83e-5 & & & \\
30 &20 &10 &4.97e-5 & & & \\
30 &25 &5 &9.97e-6 & & & \\
\specialrule{0.3pt}{0.1pt}{0.1pt}
\textbf{30} &\textbf{27} &\textbf{3} &\textbf{8.88e-6} & & & \\
40 &0 &40 &0.002634 &\multirow{12}{*}{0.77} &\multirow{12}{*}{0.13} &\multirow{12}{*}{0.14} \\
40 &2 &38 &2.69e-4 & & & \\
40 &4 &36 &6.84e-6 & & & \\
40 &5 &35 &1.38e-4 & & & \\
40 &8 &32 &7.96e-5 & & & \\
40 &10 &30 &3.79e-5 & & & \\
40 &15 &25 &2.26e-5 & & & \\
40 &16 &24 &9.11e-6 & & & \\
40 &20 &20 &6.41e-6 & & & \\
\textbf{40} &\textbf{25} &\textbf{15} &\textbf{2.25e-6} & & & \\
40 &30 &10 &7.04e-6 & & & \\
40 &35 &5 &2.02e-5 & & & \\
\bottomrule
\end{tabular}
\end{threeparttable}
}
\end{adjustwidth}

%% file: supp_tables/div2k-dr-m1.tex
\begin{adjustwidth}{-2.5 cm}{-2.5 cm}\centering
\resizebox{\columnwidth}{!}{
\begin{threeparttable}[p]
\caption{DIV2k: $\Lambda_{M_1}$}\label{tab:div2k-dr-m1}
\scriptsize
\begin{tabular}{lrrrrrrr}\toprule\toprule
\multirow{2}{*}{\textbf{Target Dimension}} &\multirow{2}{*}{\textbf{k1}} &\multirow{2}{*}{\textbf{k2}} &\multirow{2}{*}{\textbf{M1}} &\multirow{2}{*}{\textbf{PCA M1}} &\multicolumn{2}{c}{\textbf{RMap M1 ($\mathbf{\alpha}$=20)}} \\\cmidrule{6-7}
& & & & &$\mathbf{\mu}$ &$\mathbf{\sigma}$ \\\midrule
10 &0 &10 &0.001538 &\multirow{8}{*}{0.66} &\multirow{8}{*}{0.05} &\multirow{8}{*}{0.029} \\
10 &1 &9 &0.00092 & & & \\
10 &2 &8 &0.000219 & & & \\
10 &3 &7 &0.000219 & & & \\
10 &4 &6 &0.0007 & & & \\
10 &5 &5 &0.000289 & & & \\
\textbf{10} &\textbf{6} &\textbf{4} &\textbf{7.07e-5} & & & \\
10 &7 &3 &0.000231 & & & \\
\textbf{20} &\textbf{0} &\textbf{20} &\textbf{3.01e-6} &\multirow{10}{*}{0.58} &\multirow{10}{*}{0.04} &\multirow{10}{*}{0.027} \\
\specialrule{0.3pt}{0.1pt}{0.1pt}
20 &2 &18 &7.61e-5 & & & \\
20 &3 &17 &9.13e-5 & & & \\
20 &4 &16 &3.25e-5 & & & \\
20 &5 &15 &9.80e-5 & & & \\
20 &8 &12 &0.000158 & & & \\
20 &10 &10 &0.000147 & & & \\
20 &12 &8 &9.74e-5 & & & \\
20 &15 &5 &4.49e-5 & & & \\
20 &18 &2 &0.00063 & & & \\
\specialrule{0.3pt}{0.1pt}{0.1pt}
30 &0 &30 &0.000279 &\multirow{12}{*}{0.54} &\multirow{12}{*}{0.02} &\multirow{12}{*}{0.020} \\
30 &2 &28 &0.000965 & & & \\
30 &3 &27 &8.80e-5 & & & \\
30 &5 &25 &0.000117 & & & \\
30 &8 &22 &0.00059 & & & \\
30 &10 &20 &0.000402 & & & \\
30 &12 &18 &0.000163 & & & \\
30 &15 &15 &0.000301 & & & \\
\textbf{30} &\textbf{18} &\textbf{12} &\textbf{7.46e-5} & & & \\
30 &20 &10 &0.000432 & & & \\
30 &25 &5 &0.000187 & & & \\
30 &27 &3 &0.000164 & & & \\
\specialrule{0.3pt}{0.1pt}{0.1pt}
40 &0 &40 &0.000696 &\multirow{12}{*}{0.51} &\multirow{12}{*}{0.03} &\multirow{12}{*}{0.018} \\
40 &2 &38 &0.000442 & & & \\
40 &4 &36 &5.37e-5 & & & \\
40 &5 &35 &9.36e-5 & & & \\
40 &8 &32 &6.96e-5 & & & \\
40 &10 &30 &0.00034 & & & \\
40 &15 &25 &8.92e-5 & & & \\
40 &16 &24 &6.33e-5 & & & \\
40 &20 &20 &8.28e-6 & & & \\
40 &25 &15 &0.000162 & & & \\
\textbf{40} &\textbf{30} &\textbf{10} &\textbf{4.00e-6} & & & \\
40 &35 &5 &0.000191 & & & \\
\bottomrule
\end{tabular}
\end{threeparttable}
}
\end{adjustwidth}